\documentclass[journal,11pt]{IEEEtran}
\usepackage{cite}
\usepackage[pdftex]{graphicx}
\usepackage[cmex10]{amsmath}
\usepackage[ruled,vlined]{algorithm2e}
\usepackage{array}
\usepackage{fixltx2e}
\usepackage{color}
\usepackage{multirow}
\usepackage[normalem]{ulem}
\usepackage[percent]{overpic}

\graphicspath{{./}}

\newcommand {\ADD}[1]{#1}

\newcommand{\coso}{strategy}
\newcommand{\emb}{embodiment}

\newcommand{\curlef}{CURL-EF(EP+LR)}
\newcommand{\curllf}{CURL-LF(EP+LR)}
\newcommand{\curlefn}{CURL-EF$_n$(EP+LR)}
\newcommand{\curllfn}{CURL-LF$_n$(EP+LR)}
\newcommand{\curleflf}{CURL-EF\&LF(EP+LR)}
\newcommand{\curleflfn}{CURL-EF\&LF$_n$(EP+LR)}
\newcommand{\curllaplf}{CURL-LF(LapSVM)}
\newcommand{\curllaplfn}{CURL-LF$_n$(LapSVM)}
\newcommand{\curllapef}{CURL-EF(LapSVM)}
\newcommand{\curllapefn}{CURL-EF$_n$(LapSVM)}

\begin{document}

\title{CURL: Co-trained Unsupervised Representation Learning for Image Classification}

\author{Simone~Bianco,
        Gianluigi~Ciocca,
        and~Claudio~Cusano
\thanks{S. Bianco, and G. Ciocca are with the Department
of Informatics Systems and Communications, University of Milano-Bicocca, Milano,
20126 Italy, e-mail: \{bianco, ciocca\}@disco.unimib.it.}
\thanks{C. Cusano is with the Dipartimento di Ingegneria Industriale e dell'Informazione, University of Pavia, Pavia, 27100 Italy. e-mail: claudio.cusano@unipv.it}
}


\maketitle

\begin{abstract}
In this paper we propose a \coso{ }for semi-supervised image classification that leverages unsupervised representation learning and co-training. The \coso, that is called CURL from Co-trained Unsupervised Representation Learning, iteratively builds two classifiers on two different views of the data. 
The two views correspond to different representations learned from both labeled and unlabeled data and differ in the fusion scheme used to combine the image features. 

To assess the performance of our proposal, we conducted several experiments on widely used data sets for scene and object recognition.  We considered three scenarios (inductive, transductive  and self-taught learning) that differ in the strategy followed to exploit the unlabeled data. \ADD{As image features we considered a combination of GIST, PHOG, and LBP as well as features extracted from a Convolutional Neural Network. Moreover, two \emb{s} of CURL are investigated: one using Ensemble Projection as unsupervised representation learning coupled with Logistic Regression, and one based on LapSVM.} The results show that CURL clearly outperforms other supervised and semi-supervised learning methods in the state of the art.

\end{abstract}


\section{Introduction}
\label{sec:intro}

Semi-supervised learning~\cite{chapelle2006semi} consists in taking
into account both labeled and unlabeled data when training machine
learning models.  It is particularly effective when there is plenty of
training data, but only a few instances are labeled.  In the last
years, many semi-supervised learning approaches have been proposed
including generative methods~\cite{nigam2000text,fujino2005hybrid},
graph-based methods~\cite{blum2001learning,chapelle2002cluster}, and
methods based on Support Vector
Machines~\cite{joachims1999transductive,belkin2006manifold}.
Co-training is another example of semi-supervised
technique~\cite{blum1998combining}.  It consists in training two
classifiers independently which, on the basis of their level of
confidence on unlabeled data, co-train each other trough the
identification of good additional training examples.  The difference
between the two classifiers is that they work on different
\emph{views} of the training data, often corresponding to two feature
vectors.  Pioneering works on co-training identified the conditional
independence between the views as the main reason of its success.
More recently, it has been observed that conditional independence is a
sufficient, but not necessary condition, and that even a single view
can be considered, provided that different classification techniques
are used~\cite{zhou2010semi}.

\begin{figure}%
	\includegraphics[width=\columnwidth]{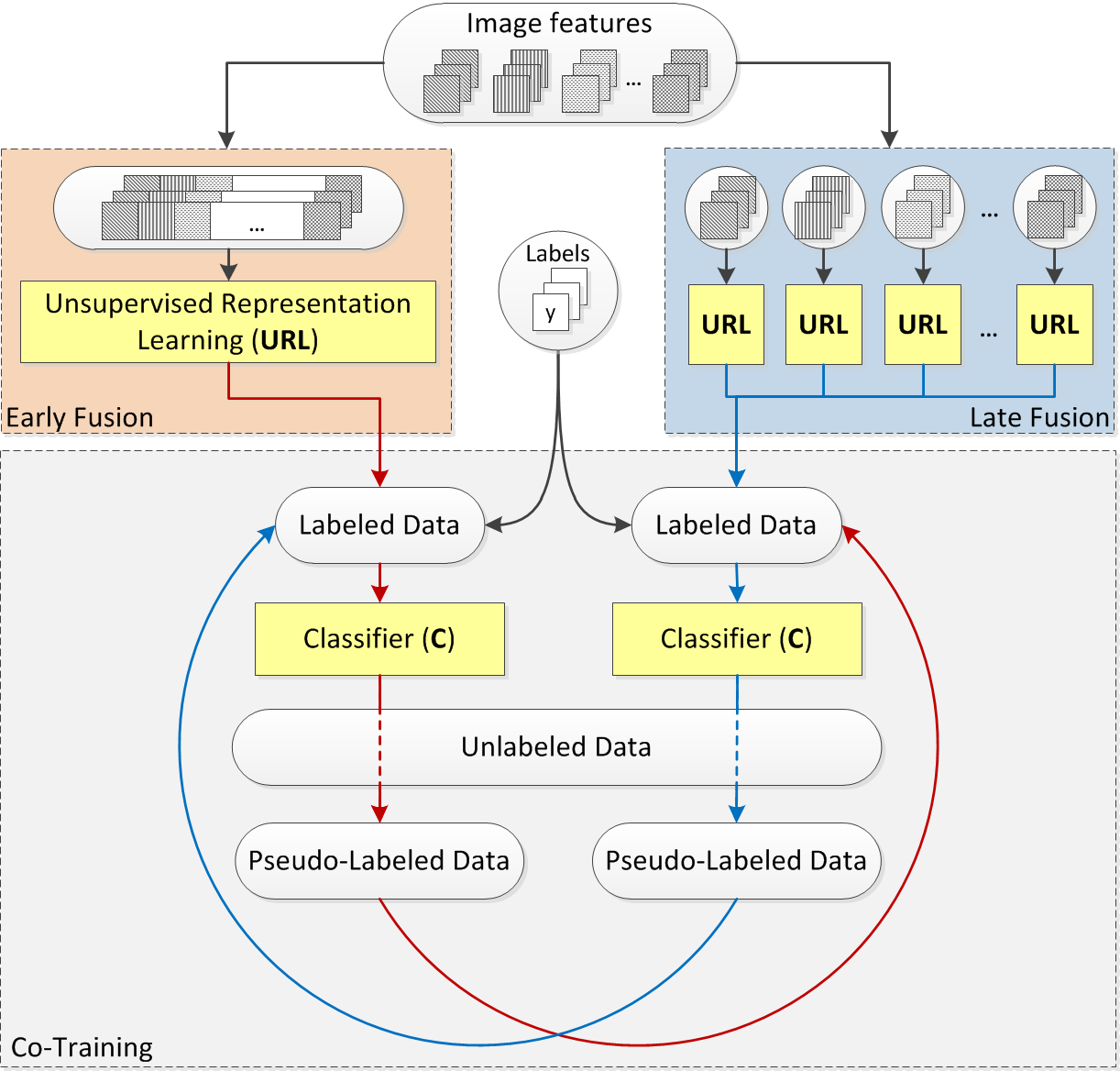}%
  \caption{Schema of the proposed \coso.}%
  \label{fig:schema}%
\end{figure}

In this work we propose a semi-supervised image classification
strategy which exploits unlabeled data in two different ways: first
two image representations are obtained by unsupervised representation learning \ADD{(URL)} on a
set of image features computed on all the available training data;
then co-training is used to enlarge the labeled training set of the
corresponding co-trained classifiers \ADD{(C)}.  The difference between the two image
representations is that one is built on the combination of all the
image features (\emph{early fusion}), while the other is the
combination of sub-representations separately built on each feature 
(\emph{late fusion}).  \ADD{We call the proposed strategy CURL: Co-trained Unsupervised Representation Learning  (from the combination of C and URL components). The schema of CURL} is illustrated in
Fig.~\ref{fig:schema}.

In standard co-training each classifier is built on a single view,
often corresponding to a single feature.  However, the combination of
multiple features is often required to recognize complex visual
concepts~\cite{Iyengar03,Gehler09,natarajan12}. Both the classifiers built by CURL exploit all
the available image features in such a way that these concepts can be
accurately recognized.  \ADD{We argue that} the use of two different fusion schemes
together with the non-linear transformation produced by the
unsupervised learning procedure, makes the two image representations
uncorrelated enough to allow an effective co-training of the
classifiers.

\ADD{The proposed \coso{ }is built on two base components: URL (the unsupervised representation learning) and C (the classifier used in co-training). 
By changing these two components we can have different \emb{s} of CURL that can be experimented and evaluated.}

To assess the merits of our proposal we conducted several experiments
on widely used data sets: the 15-scene data set\ADD{, the
Caltech-101 object classification data set, and the ILSVCR 2012 data set which contains 1000 different classes}.  We considered a variety
of scenarios including \emph{transductive learning} (i.e.~unlabeled test data
available during training), \emph{inductive learning} (i.e.~test data not
available during training), and \emph{self-taught learning} (i.e.~test and
training data coming from two different data sets).  \ADD{In order to verify the efficacy of the CURL classification \coso, we also tested two \emb{s}: one that uses Ensemble Projection unsupervised representation coupled with Logistic Regression classification, and one based on LapSVM semi-supervised classification.
Moreover different variants of the \emb{s} are evaluated as well.} The results
show that CURL clearly outperforms other semi-supervised learning
methods in the state of the art.


\section{Related Work}
\label{sec:related}

There is a large literature on semi-supervised learning. For the sake
of brevity, we discuss only the paradigms involved in the proposed
\coso.  More information about these and other approaches to
semi-supervised learning can be found in the book by Chapelle \emph{et
  al.}~\cite{chapelle2006semi}.
	
\subsection{Co-training}
Blum and Mitchell proposed co-training in
1998~\cite{blum1998combining} and verified its effectiveness for the
classification of web pages.  The basic idea is that two classifiers
are trained on separate views (features) and then used to train each
other.  More precisely, when one of the classifiers is very confident
in making a prediction for unlabeled data, the predicted labels are
used to augment the training set of the other classifier.  
The concept has been generalized to three \cite{zhou2005tritraining} or more views \cite{li2007improve,zhou2011semi}. 
Co-training has been used in several computer vision applications including video
annotation~\cite{wang2006enhanced}, action
recognition~\cite{gupta2008watch}, traffic
analysis~\cite{levin2003unsupervised}, speech and gesture
recognition~\cite{christoudias2006coadaptation}, image
annotation~\cite{feng2003bootstrapping}, biometric
recognition~\cite{bhatt2011cotraining}, image
retrieval~\cite{tong2001support}, image
classification~\cite{guillaumin2010multimodal}, 
object detection~\cite{levin2003unsupervised,javed2005online}, and object tracking~\cite{tang2007co}. 

According to Blum and Mitchell, a sufficient condition for the
effectiveness of co-training is that, beside being individually
accurate, the two classifiers are conditionally independent given the
class label.  However, conditional independence is not a necessary
condition.  In fact, Whang and Zhou~\cite{wang2007analyzing} showed
that co-training can be effective when the diversity between the two
classifiers is larger than their errors; their results provided a
theoretical support to the success of single-view co-training
variants~\cite{goldman2000enhancing,chen2011automatic,wang2013co} (the reader
may refer to an updated study from the same authors~\cite{wang2010new}
for more details about necessary and sufficient conditions for
co-training).

\subsection{Unsupervised representation learning}
In the last years,
as a consequence of the success of deep learning frameworks we
observed an increased interest in methods that make use of unlabeled
data to automatically learn new representations.  In fact, these have
been demonstrated to be very effective for the pre-training of large
neural networks~\cite{hinton2006fast,jarrett2009best}.  Restricted
Boltzmann Machines~\cite{hinton2002training} and auto-encoder
networks~\cite{bourlard1988autoassociation} are notable examples of
this kind of methods.  The tutorial by Bengio covers in detail this
family of approaches~\cite{bengio2009learning}.

A conceptually simpler approach consists in using clustering
algorithms to identify frequently occurring patterns in unlabeled data
that can be used to define effective representations.  The K-means
algorithm has been widely used for this
purpose~\cite{coates2012learning}.  In computer vision this approach
is very popular and lead to the many variants of bag-of-visual-words
representations~\cite{csurka2004visual,lazebnik2006beyond}.  Briefly,
clustering on unlabeled data is used to build a vocabulary of visual
words.  Given an image, multiple local features are extracted and for
each of them the most similar visual word is searched.  The final
representation is a histogram counting the occurrences of the visual
words.  Sparse coding can be seen as an extension of this approach,
where each local feature is described as a sparse combination of
multiple words of the
vocabulary~\cite{olshausen1997sparse,mairal2010online,lewicki2000learning}.

Another strategy for unsupervised feature learning is represented by
Ensemble Projection (EP)~\cite{dai2013ensemble}. From all the
available data (labeled and unlabeled) Ensemble Projection samples a
set of prototypes. Discriminative learning is then used to learn
projection functions tuned to the prototypes.  
Since a single set of projections could be too noisy, multiple sets of
prototypes are sampled to build an ensemble of projection functions. The values computed according to these functions represent the
components of the learned representations.

LapSVM \cite{belkin2006manifold} can be seen as an unsupervised representation learning method as well. In this case the learned representation is not explicit but it is implicitly embedded in a kernel learned from unlabeled data. 

\subsection{Fusion schemes}
\ADD{
Combining multimodal information is an important issue in pattern recognition. The 
fusion of multimodal inputs can bring complementary information from various
sources, useful for improving the quality of the image retrieval and classification performance \cite{atrey2010multimodal}. The problem arises in defining how these modalities are to be 
combined or fused. In general, 
the existing fusion approaches can be categorized as early and late fusion
approaches, which refers to their relative position from the
feature comparison or learning step in the whole processing chain. Early fusion usually refers 
to the combination of the features into a single representation before comparison/learning. 
Late fusion refers to the combination, at the last stage, of the responses obtained after 
individual features comparison or learning \cite{snoek2005early,noble04}.
There is no universal conclusion as to which strategy is
the preferred method for for a given task. For example, Snoek et al. \cite{snoek2005early}
found that late fusion is better than early fusion in the TRECVID 2004 semantic
indexing task, while Ayache et al. \cite{ayache2007classifier} stated that early fusion gets better results
than late fusion on the TRECVID 2006 semantic indexing task.
A combination of these approaches can also be exploited as hybrid fusion
approach \cite{wu2005multi}.
}

\ADD{Another form of data fusion is Multiple Kernel Learning (MKL). MKL has been introduced by Lanckriet et al. \cite{lanckriet04} as extension of the support vector machines (SVMs). Instead of using a single kernel computed on the image representation as in standard SVMs, MKL learns distinct kernels. The kernels are combined with a linear or non linear function and the function's parameters can be determined during the learning process. MKL can be used to learn different kernels on the same image representation or by learning different kernels each one on a different image representation \cite{gonen2011multiple}. The former corresponds to have different notion of similarity, and to choose the most suitable one for the problem and representation at hand. The latter corresponds to have multiple representations each with a, possibly, different definition of similarity that must be combined together. This kind of data fusion, in \cite{noble04}, is termed \emph{intermediate fusion}.}

\section{The Proposed \coso: CURL}
\label{sec:curl}
In the semi-supervised image classification setup the training data consists of both labeled examples $\{\mathcal{X}_l,\mathcal{Y} \}=\{ ({\mathbf{x}}_i,y_i) \}_{\scriptscriptstyle i=1}^{\scriptscriptstyle L}$ and unlabeled ones $\mathcal{X}_u=\{ {\mathbf{x}}_i \}_{\scriptscriptstyle i=L+1}^{\scriptscriptstyle L+U}$, where $\mathbf{x}_i$ denotes the feature vector of image $i$, $y_i \in \{1, \ldots, K\}$ is its label, and $K$ is the number of classes. 


In this work, for each image $i$ a set of $S$ different image features $\mathbf{x}_i^{(s)}$, $s=1,\ldots,S$ is considered. Two views are then generated by using two different fusion strategies: early and late fusion. 
In case of Early Fusion (EF), the image features are concatenated 
and then 
used to learn a new representation $\mathbf{x}_i^{\scriptscriptstyle EF}=\varphi(\{[\mathbf{x}_i^{(1)},\ldots,\mathbf{x}_i^{(S)}]\}$ in an unsupervised way, where $\varphi (\cdot)$ is a projection function.  
In case of Late Fusion (LF), an unsupervised representation $\varphi_s(\mathbf{x}_i^{(s)})$ is independently learned for each image feature and then the representations are concatenated to obtain $\mathbf{x}_i^{\scriptscriptstyle LF}=[\varphi_1(\mathbf{x}_i^{(1)}),\ldots,\varphi_S(\mathbf{x}_i^{(S)})]$.

Using the learned EF and LF unsupervised representations, the two views are built: 
$\mathcal{X}_l^{\scriptscriptstyle EF}=\{\mathbf{x}_i^{\scriptscriptstyle EF}\}_{\scriptscriptstyle i=1}^{\scriptscriptstyle L}$, 
$\mathcal{X}_u^{\scriptscriptstyle EF}=\{\mathbf{x}_i^{\scriptscriptstyle EF}\}_{\scriptscriptstyle i=L+1}^{\scriptscriptstyle L+U}$ and  
$\mathcal{X}_l^{\scriptscriptstyle LF}=\{\mathbf{x}_i^{\scriptscriptstyle LF}\}_{\scriptscriptstyle i=1}^{\scriptscriptstyle L}$, 
$\mathcal{X}_u^{\scriptscriptstyle LF}=\{\mathbf{x}_i^{\scriptscriptstyle LF}\}_{\scriptscriptstyle i=L+1}^{\scriptscriptstyle L+U}$. 
Furthermore, two label sets $\mathcal{Y}^{\scriptscriptstyle EF}$ and $\mathcal{Y}^{\scriptscriptstyle LF}$ are initialized equal to $\mathcal{Y}$.

Once the two views are generated, our method iteratively co-trains two classifiers $\phi_{\scriptscriptstyle EF}$ and $\phi_{\scriptscriptstyle LF}$ on them \cite{blum1998combining}. SVMs, logistic regressions, or any other similar technique can be used to obtain them. 
The idea of iterative co-training is that one can use a small labeled sample to train the initial classifiers over the respective views (i.e. $\phi_{\scriptscriptstyle EF}: \mathcal{X}_l^{\scriptscriptstyle EF} \mapsto {\mathcal{Y}}^{\scriptscriptstyle EF}$ and $\phi_{\scriptscriptstyle LF}: \mathcal{X}_l^{\scriptscriptstyle LF} \mapsto {\mathcal{Y}}^{\scriptscriptstyle LF}$), and then iteratively bootstrap by taking unlabeled examples for which one of the classifiers is confident but the
other is not. The confident classifier determines pseudo-labels \cite{lee2013pseudo} that are then used as if they were true labels to improve the other classifier \cite{balcan2004co}. 
Given the classifier confidence scores $\mathbf{w}_i^{\scriptscriptstyle EF}=\phi_{\scriptscriptstyle EF}(\mathbf{x}_i^{\scriptscriptstyle EF})$ and $\mathbf{w}_i^{\scriptscriptstyle LF}=\phi_{\scriptscriptstyle LF}(\mathbf{x}_i^{\scriptscriptstyle LF})$, the pseudo-labels $\hat{y}_i^{\scriptscriptstyle EF}$ and $\hat{y}_i^{\scriptscriptstyle LF}$ are respectively obtained as:
\begin{equation}
\hat{y}_i^{\scriptscriptstyle EF}={\text{arg}} \!\!\! \max_{j=1,\ldots,K} \mathbf{w}_i^{\scriptscriptstyle EF}[j]
\label{eq:pseudo1}
\end{equation}
\begin{equation}
\hat{y}_i^{\scriptscriptstyle LF}={\text{arg}} \!\!\! \max_{j=1,\ldots,K} \mathbf{w}_i^{\scriptscriptstyle LF}[j]
\label{eq:pseudo2}
\end{equation}

In each round of co-training, the classifier $\phi_{\scriptscriptstyle LF}$ chooses some examples in $\mathcal{X}_u^{\scriptscriptstyle EF}$ to pseudo-label for $\phi_{\scriptscriptstyle EF}$, and vice versa. 
For each class $k$, let us call $\mathcal{X}_\star$ the set of candidate unlabeled examples to be pseudo-labeled for $\phi_{\scriptscriptstyle EF}$. Each $\mathbf{x}_\star \in \mathcal{X}_\star$ must belong to the unlabeled set, i.e. $\mathbf{x}_\star \in \mathcal{X}_u^{\scriptscriptstyle EF}$, has not to be already used for training, i.e. $\mathbf{x}_\star \notin \mathcal{X}_l^{\scriptscriptstyle EF}$, and its pseudo-label has to be $\hat{y}_\star^{\scriptscriptstyle LF}=k$. Furthermore, $\phi_{\scriptscriptstyle LF}$ should be more confident on the classification of $\mathbf{x}_\star$ than $\phi_{\scriptscriptstyle EF}$, and its confidence should be higher than a fixed threshold $t_1$:

\begin{equation}
\forall \mathbf{x}_\star \in \mathcal{X}_\star: \mathbf{w}_\star^{\scriptscriptstyle EF}[k]<\mathbf{w}_\star^{\scriptscriptstyle LF}[k], \mathbf{w}_\star^{\scriptscriptstyle LF}[k]  > \! t_1
\label{eq:updateEF}
\end{equation}

If no $\mathbf{x}_\star$ satisfying Eq. \ref{eq:updateEF} are found, then the constraints are relaxed:
\begin{equation}
\forall \mathbf{x}_\star \in \mathcal{X}_\star: \mathbf{w}_\star^{\scriptscriptstyle LF}[k]  > \! t_2, \text{ with } t_2<t_1
\label{eq:updateEF2}
\end{equation}

Non-maximum suppression is applied to add one single pseudo-labeled example for each class by extracting the most confident $\mathbf{x}_\star \in \mathcal{X}_\star$:

\begin{equation}
\text{find } \mathbf{x}_\star  \in  \mathcal{X}_\star : \mathbf{w}_\star^{\scriptscriptstyle LF}[k]  = \displaystyle \text{arg} \! \max_{j} \mathbf{w}_j^{\scriptscriptstyle LF}[k] 
\label{eq:coso}
\end{equation}
The selected $\mathbf{x}_\star$ and its corresponding pseudo-label $\hat{y}_\star$ are added to $\mathcal{X}_l^{\scriptscriptstyle LF}$ and $\mathcal{Y}^{\scriptscriptstyle LF}$ respectively. If no $\mathbf{x}_\star$ satisfying Eq. \ref{eq:updateEF2} are found, then nothing is added to $\mathcal{X}_l^{\scriptscriptstyle LF}$ and $\mathcal{Y}^{\scriptscriptstyle LF}$.

Similarly, the classifier $\phi_{\scriptscriptstyle EF}$ chooses some examples in $\mathcal{X}_u^{\scriptscriptstyle LF}$ to pseudo-label for $\phi_{\scriptscriptstyle LF}$. 
At the next co-training round, two new classifiers $\phi_{\scriptscriptstyle EF}$ and $\phi_{\scriptscriptstyle LF}$ are trained on the respective views, that now contain both labeled and pseudo-labeled examples. 
The complete procedure of the CURL method is outlined in Algorithms \ref{alg:curl}-\ref{alg:nms}.

\setlength{\algomargin}{-0.005em}
\begin{algorithm}[!tbp]
\KwData{Labeled data $\{\mathcal{X}_l,\mathcal{Y}\}$, unlabeled data $\mathcal{X}_u$}
\KwResult{Classifiers $\phi_{\scriptscriptstyle EF}(\cdot)$ and $\phi_{\scriptscriptstyle LF}(\cdot)$}
\Begin{
$[\mathcal{X}_l^{\scriptscriptstyle EF} \! , \! \mathcal{X}_u^{\scriptscriptstyle EF} \! , \! \mathcal{X}_l^{\scriptscriptstyle LF} \! , \! \mathcal{X}_u^{\scriptscriptstyle LF}]=\text{\bf{computeURL}}(\mathcal{X}_l,\mathcal{X}_u)$\\
${\mathcal{Y}}^{\scriptscriptstyle EF}={\mathcal{Y}}^{\scriptscriptstyle LF}={\mathcal{Y}}$ \\
train classifier $\phi_{\scriptscriptstyle EF}: \mathcal{X}_l^{\scriptscriptstyle EF} \mapsto {\mathcal{Y}}^{\scriptscriptstyle EF}$\\
train classifier $\phi_{\scriptscriptstyle LF}: \mathcal{X}_l^{\scriptscriptstyle LF} \mapsto {\mathcal{Y}}^{\scriptscriptstyle LF}$\\
\For{co-training round $c=1:C$}{
	  initialize $\mathcal{W}^{\scriptscriptstyle EF}=\mathcal{W}^{\scriptscriptstyle LF}=\hat{\mathcal{Y}}^{\scriptscriptstyle EF}=\hat{\mathcal{Y}}^{\scriptscriptstyle LF}=\emptyset$\\

	  \ForEach{$\mathbf{x}_i^{\scriptscriptstyle EF} \in \mathcal{X}_u^{\scriptscriptstyle EF}$}{
				add $\mathbf{w}_i^{\scriptscriptstyle EF}=\phi_{\scriptscriptstyle EF}(\mathbf{x}_i^{\scriptscriptstyle EF})$ to $\mathcal{W}^{\scriptscriptstyle EF}$\\
				add $\displaystyle \hat{y}_i^{\scriptscriptstyle EF}={\text{arg}} \!\!\! \max_{j=1,\ldots,K} \mathbf{w}_i^{\scriptscriptstyle EF}[j]$ to $\hat{\mathcal{Y}}^{\scriptscriptstyle EF}$\\
				}
		\ForEach{$\mathbf{x}_i^{\scriptscriptstyle LF} \in \mathcal{X}_u^{\scriptscriptstyle LF}$}{
				add $\mathbf{w}_i^{\scriptscriptstyle LF}=\phi_{\scriptscriptstyle LF}(\mathbf{x}_i^{\scriptscriptstyle LF})$  to $\mathcal{W}^{\scriptscriptstyle LF}$\\
				add $\displaystyle \hat{y}_i^{\scriptscriptstyle LF}={\text{arg}} \!\!\! \max_{j=1,\ldots,K} \mathbf{w}_i^{\scriptscriptstyle LF}[j]$ to $\hat{\mathcal{Y}}^{\scriptscriptstyle LF}$\\
				}
	  
		\For{class number $k=1:K$}{

				\For{$(v_1,v_2) \in \{(\text{\footnotesize{EF},\footnotesize{LF}}),(\text{\footnotesize{LF},\footnotesize{EF}}) \}$}{
				
				find $\{\mathcal{X}_\star,\hat{\mathcal{Y}}_\star^{v_2} \}$ with \\
				$\mathcal{X}_\star \! \subset \! \mathcal{X}_u^{v_1}$, $\mathcal{X}_\star \! \cap \! \mathcal{X}_l^{v_1} \! = \! \emptyset$, $\hat{\mathcal{Y}}_\star^{v_2} \subset \hat{\mathcal{Y}}^{v_2}$ s.t.: \\
				$\forall \mathbf{x}_\star \! \in \! \mathcal{X}_\star$ and $\forall \hat{y}_\star^{v_2} \! \in \! \hat{\mathcal{Y}}_\star^{v_2}$, hold: \\ 
				$\hat{y}_\star^{v_2}=k, \mathbf{w}_\star^{v_1}[k]<\mathbf{w}_\star^{v_2}[k], \mathbf{w}_\star^{v_2}[k]  > \! t_1$\\
				\If{$  \mathcal{X}_\star = \emptyset$}{
						find $\{\mathcal{X}_\star,\hat{\mathcal{Y}}_\star^{v_2} \}$ with \\
						$\mathcal{X}_\star \! \subset \! \mathcal{X}_u^{v_1}$, $\mathcal{X}_\star \! \cap \! \mathcal{X}_l^{v_1} \! = \! \emptyset$, $\hat{\mathcal{Y}}_\star^{v_2} \! \subset \! \hat{\mathcal{Y}}^{v_2}$ s.t.: \\
						$\forall \mathbf{x}_\star \! \in \! \mathcal{X}_\star$ and $\forall \hat{y}_\star^{v_2} \! \in \! \hat{\mathcal{Y}}_\star^{v_2}$, hold: \\ 
						$\hat{y}_\star^{v_2}=k$, $\mathbf{w}_\star^{v_2}[k]>t_2$\\
						}

				$[\mathcal{X}_\star,\! \hat{\mathcal{Y}}_\star^{v_2} \!] \! = \! \text{\bf{nonMaxSuppr}}(\! \mathcal{X}_\star ,\! \hat{\mathcal{Y}}_\star^{v_2} ,\! \mathcal{W}^{v_2} \! )$ \\
				$\mathcal{X}_l^{v_1}=\mathcal{X}_l^{v_1} \cup {\mathcal{X}}_\star$ \\
				$\mathcal{Y}^{v_1}=\mathcal{Y}^{v_1} \cup \hat{\mathcal{Y}}_\star^{v_2}$ \\
				
						}
						
				}

			train classifier $\phi_{\scriptscriptstyle EF}: \mathcal{X}_l^{\scriptscriptstyle EF} \mapsto {\mathcal{Y}}^{\scriptscriptstyle EF}$\\
			train classifier $\phi_{\scriptscriptstyle LF}: \mathcal{X}_l^{\scriptscriptstyle LF} \mapsto {\mathcal{Y}}^{\scriptscriptstyle LF}$\\					
		}

} 
\label{alg:curl}
\caption{CURL}
\end{algorithm}

\begin{algorithm}[!tbp]
\KwData{Labeled data $\mathcal{X}_l$ and unlabeled data $\mathcal{X}_u$}
\KwResult{Unsup. representations $\mathcal{X}_l^{EF},\mathcal{X}_u^{EF},\mathcal{X}_l^{LF},\mathcal{X}_u^{LF}$}
\Begin{


Learn EF representation $\varphi$ on $\{[\mathbf{x}_i^{(1)},\ldots,\mathbf{x}_i^{(S)}]\}_{i=1}^{L+U}$\\
$\mathcal{X}_l^{\scriptscriptstyle EF}=\varphi(\{[\mathbf{x}_i^{(1)},\ldots,\mathbf{x}_i^{(S)}]\}_{i=1}^{L})$\\
$\mathcal{X}_u^{\scriptscriptstyle EF}=\varphi(\{[\mathbf{x}_i^{(1)},\ldots,\mathbf{x}_i^{(S)}]\}_{i=L+1}^{L+U})$\\
Learn LF representations $\varphi_s$ on $\{\mathbf{x}_i^{(s)}\}_{i=1}^{L+U}$ 
$\mathcal{X}_l^{\scriptscriptstyle LF}=\{[\varphi_1(\mathbf{x}_i^{(1)}),\ldots,\varphi_S(\mathbf{x}_i^{(S)})]\}_{i=1}^{L}$\\
$\mathcal{X}_u^{\scriptscriptstyle LF}=\{[\varphi_1(\mathbf{x}_i^{(1)}),\ldots,\varphi_S(\mathbf{x}_i^{(S)})]\}_{i=L+1}^{L+U}$\\

}
\caption{compute URL}
\label{alg:url}
\end{algorithm}


\begin{algorithm}[!tbp]
\KwData{$\mathcal{X}, \hat{\mathcal{Y}}, \mathcal{W}, k$}
\KwResult{$\mathcal{X}_\star, \hat{\mathcal{Y}}_\star$}
\Begin{
find $\{\mathcal{X}_\star, \hat{\mathcal{Y}}_\star \}$ with $\mathcal{X}_\star \in \mathcal{X}$, $\hat{\mathcal{Y}}_\star \in \hat{\mathcal{Y}}$ s.t.: \\
$\quad \mathbf{w}_\star[k] = \displaystyle \text{arg} \! \max_{j} \mathbf{w}_j[k] $, with $\mathbf{w}_j \in \mathcal{W}$\\
}
\caption{non-maximum suppression}
\label{alg:nms}
\end{algorithm}

\section{Experiments}
\label{sec:experiments}

CURL is parametric with respect to the projection function $\varphi$ used in the unsupervised representation learning \ADD{URL}, and the supervised classification technique \ADD{C} used during to co-train $\phi_{\scriptscriptstyle EF}$ and $\phi_{\scriptscriptstyle LF}$. \ADD{As first \emb{ } of CURL,} we used Ensemble Projection \cite{dai2013ensemble} for the former and logistic regression for the latter. \ADD{Another \emb, based on LapSVM \cite{belkin2006manifold} is presented in Section \ref{subsec:curllap}.}

\subsection{Data sets}
We evaluated our method on two data sets: Scene-15 (S-15) \cite{lazebnik2006beyond}, and Caltech-101 (C-101) \cite{fei2007learning}. 
Scene-15 data set contains 4485 images divided into 15 scene categories with both indoor and outdoor environments. Each category has 200 to 400 images. 
Caltech-101 contains 8677 images divided into 101 object categories, each having 31 to 800 images. 
Furthermore, we collected a set of random images by sampling 20,000 images from the ImageNet data set \cite{deng2009imagenet} to evaluate our method on the task of self-taught image classification. 
Since the current version of ImageNet has 21841 synsets (i.e. categories) and a total of more than 14 millions images, there is a small probability that the random images and images in the two considered data sets  come from the same distribution. 

\subsection{Image features}
In our experiments we used the following three features: GIST \cite{oliva2001modeling}, Pyramid of Histogram of Oriented Gradients (PHOG) \cite{bosch2007image}, and Local Binary Patterns (LBP) \cite{ojala2002multiresolution}. 
GIST was computed on the rescaled images of 256$\times$256 pixels, at 3 scales with 4, 8 and 8 orientations respectively. 
PHOG was computed with a 2-layer pyramid and in 8 directions. 
Uniform LBP with radius equal to 1, and 8 neighbors was used.

In Section \ref{subsec:cnn} we also investigate the use of features extracted from a CNN \cite{razavian2014cnn} in combination with the previous ones. 

\subsection{Ensemble projection}
\label{sec:EP}
Differently from others semi-supervised methods that train a classifier 
from labeled data with a regularization term learned from unlabeled data, Ensemble Projection \cite{dai2013ensemble} learns a new image representation from all known data (i.e. labeled and unlabeled data), and then trains a plain classifier 
on it.

Ensemble Projection learns knowledge from $T$ different prototype sets $\mathcal{P}^t = \{ (s_i^t,c_i^t) \}_{i=1}^{rn}$, with $ t \in \{1, \ldots, T\}$ where $s_i^t \in \{1, \ldots, L+U \}$ is the index of the $i-$th chosen image, $c_i^t \in \{1, \ldots, r \}$ is the pseudo-label indicating to which prototype $s_i^t$ belong to. $r$ is the number of prototypes 
in $\mathcal{P}^t$, and $n$ is the number of images sampled for each prototype. 
For each prototype set, $m$ hypotheses are randomly sampled, and the one containing images having the largest mutual distance is kept.  

A set of discriminative classifiers $\phi^t(\cdot)$ is learned on $\mathcal{P}^t$, one for each prototype set, and the projected vectors $\phi^t(\mathbf{x}_i)$ are obtained. The final feature vector is obtained by concatenating these projected vectors. 

Following \cite{dai2013ensemble} we set $T=300$, $r=30$, $n=6$, $m=50$, using Logistic Regression (LR) as discriminative classifier $\phi^t(\cdot)$ with $C=15$. 

\ADD{Within CURL}, Ensemble Projection is used to learn both Early Fusion and Late Fusion unsupervised representations. 
In the case of Early Fusion (EF), the feature vector $\mathbf{x}_i$ is obtained concatenating the $S$ different features available $\mathbf{x}_i=[\mathbf{x}_i^{(1)},\ldots,\mathbf{x}_i^{(S)}]$, $s=1,\ldots,S$. In the case of Late Fusion (LF), the feature vector $\mathbf{x}_i$ is made by considering just one single feature at time $\mathbf{x}_i=\mathbf{x}_i^{(s)}$. 
For both EF and LF, the same number $T$ of prototypes is used in order to assure that the unsupervised representations have the same size. 

\subsection{Experimental settings}
We conducted two kinds of experiments: (1) comparison of our \coso{ }
with competing methods for semi-supervised image classification; (2)
evaluation of our method at different number of co-training rounds.
We considered three scenarios corresponding to three different ways of
using unlabeled data.  In the \emph{inductive learning} scenario 25\%
of the unlabeled data is used together with the labeled data for the
semi-supervised training of the classifier; the remaining 75\% is used
as an independent test set.  In the \emph{transductive learning}
scenario all the unlabeled data is used during both training and test.
In the \emph{self-taught learning} scenario the set of unlabeled data
is taken from an additional data set featuring a different
distribution of image content (i.e.~the 20,000 images from ImageNet);
all the unlabeled data from the original data set is used as an
independent test set.

As evaluation measure we followed \cite{dai2013ensemble} and used the multi-class average precision (MAP),
computed as the average precision over all recall values and over all
classes.  Different numbers of training images per class were tested
for both Scene-15 and Caltech-101 (i.e. 1, 2, 3, 5, 10, and 20).  All the
reported results represent the average performance over ten runs with
random labeled-unlabeled splits.

The performance of the proposed \coso{ } are compared with those of
other supervised and semi-supervised \ADD{baseline} methods. As supervised
classifiers we considered Support Vector Machines (SVM).  As
semi-supervised classifiers, we used LapSVM
\cite{sindhwani2005beyond,belkin2006manifold}. LapSVM extend the SVM
framework including a smoothness penalty term defined on the
Laplacian adjacency graph built from both labeled and unlabeled data.
For both SVM and LapSVM we experimented with the linear, RBF and
$\chi^2$ kernels computed on the concatenation of the three available
image features as in \cite{dai2013ensemble}.  The parameters of SVM and LapSVM have been determined by
a greedy search with a three-fold cross validation on the training
set. We also compared \ADD{the present \emb{ } of CURL against} Ensemble Projection coupled with a logistic regression classifier (EP+LR) as in \cite{dai2013ensemble}. 

\section{Experimental results}
\label{sec:results}
As a first experiment we compared CURL against EP+LR, and against
SVMs and LapSVMs with different kernels. Specifically, we tested the two co-trained classifiers operating on early-fused and late-fused representations, both employing EP for URL and LR as classifier C, that we call \curlef{ } and \curllf{ } respectively. 
We also included a variant of the proposed method. It differs in the number of pseudo-labeled examples that are added at each co-training round. The variant skips the non-maximum suppression step, and at each round, adds all the examples satisfying Eq. \ref{eq:updateEF}. We denote the two co-trained classifiers of the variant as \curlefn{ } and \curllfn.

\begin{figure*}[ht]%
\centering
\renewcommand{\tabcolsep}{0cm}
\begin{tabular}{ccc}
\footnotesize{Scene-15 inductive} & \footnotesize{Scene-15 transductive} & \footnotesize{Scene-15 self-taught}\\
\includegraphics[width=0.68\columnwidth]{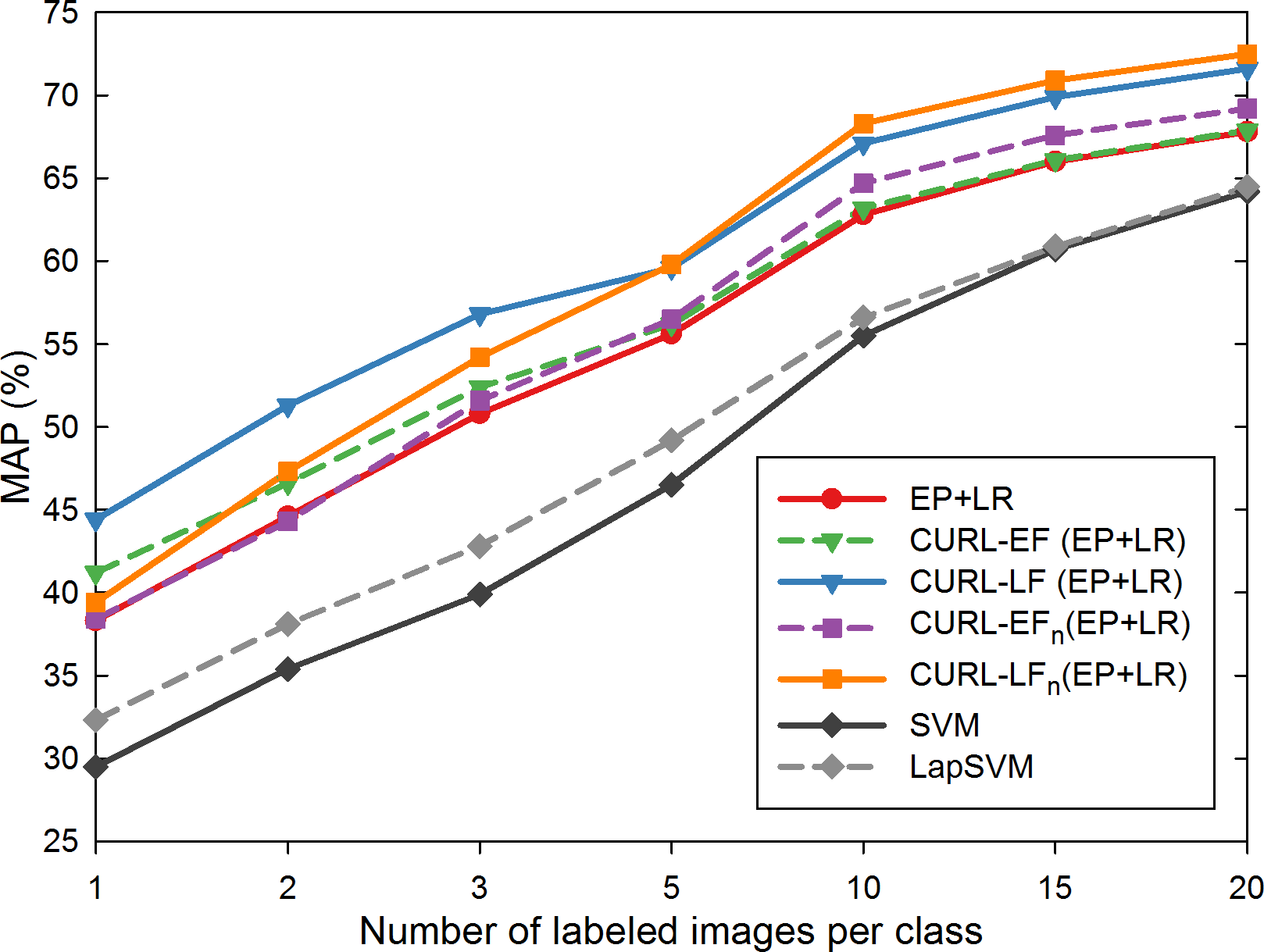} &
\includegraphics[width=0.68\columnwidth]{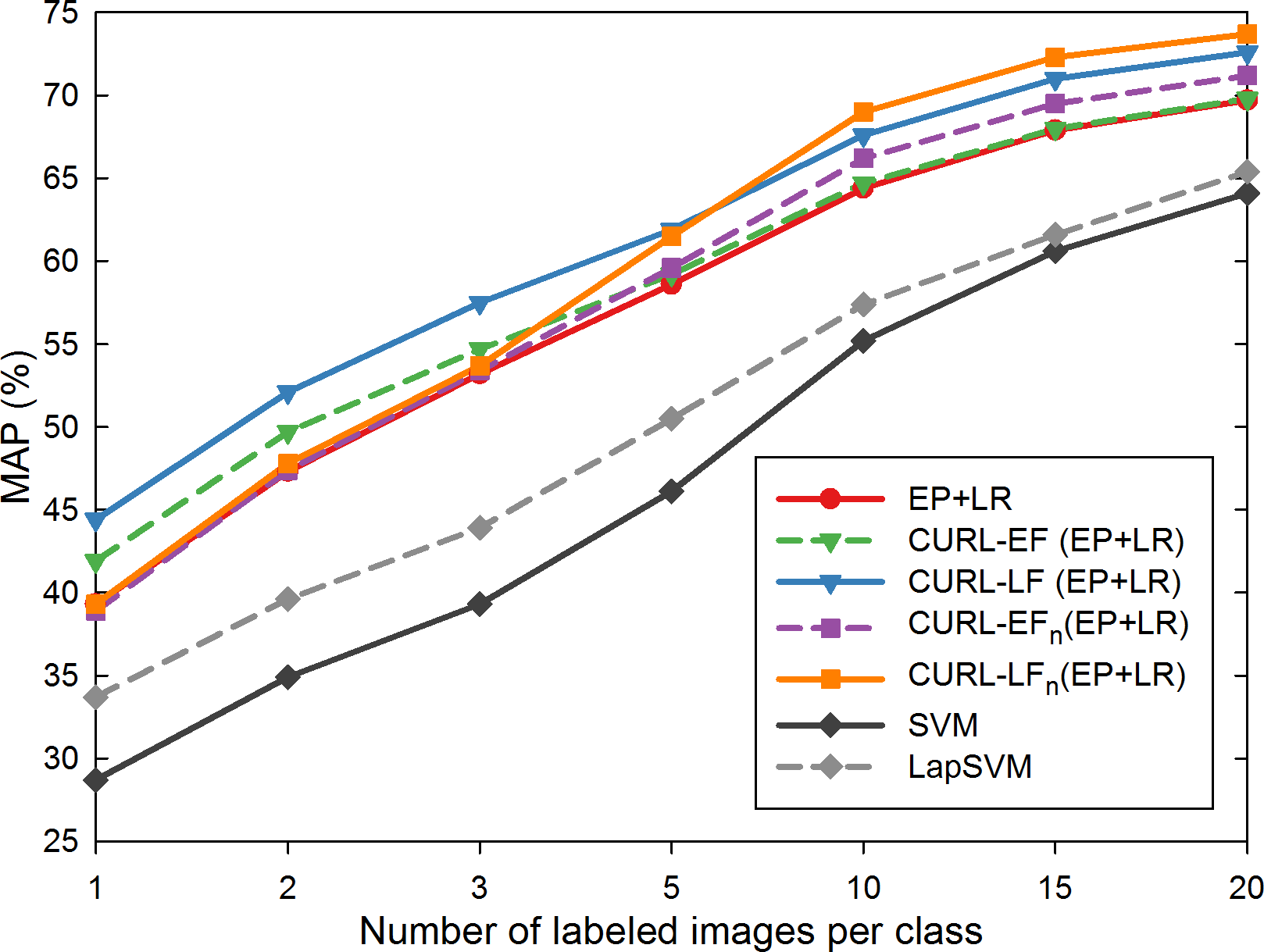} &
\includegraphics[width=0.68\columnwidth]{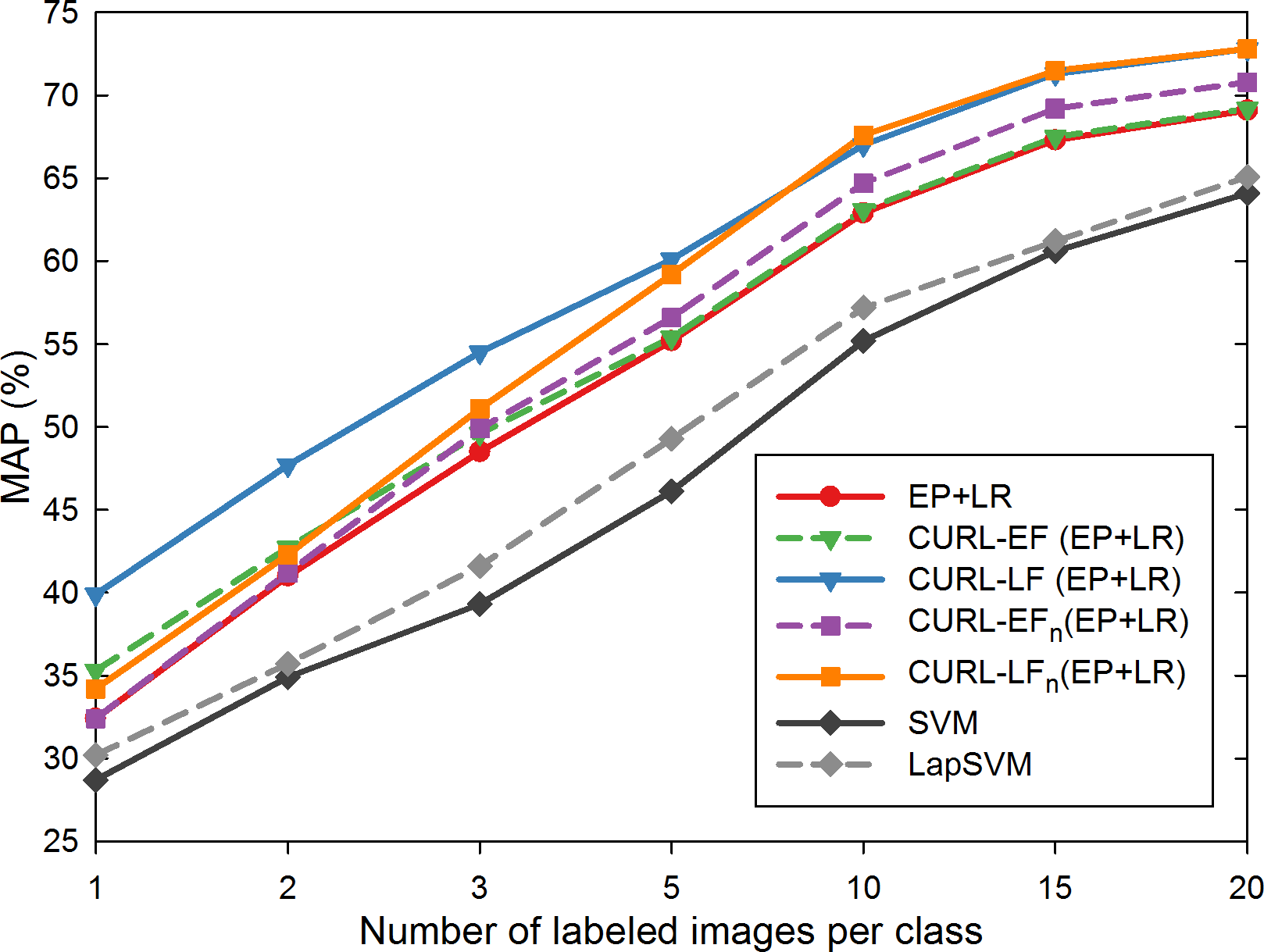} \\
 \\
\footnotesize{Caltech-101 inductive} & \footnotesize{Caltech-101 transductive} & \footnotesize{Caltech-101 self-taught}\\
\includegraphics[width=0.68\columnwidth]{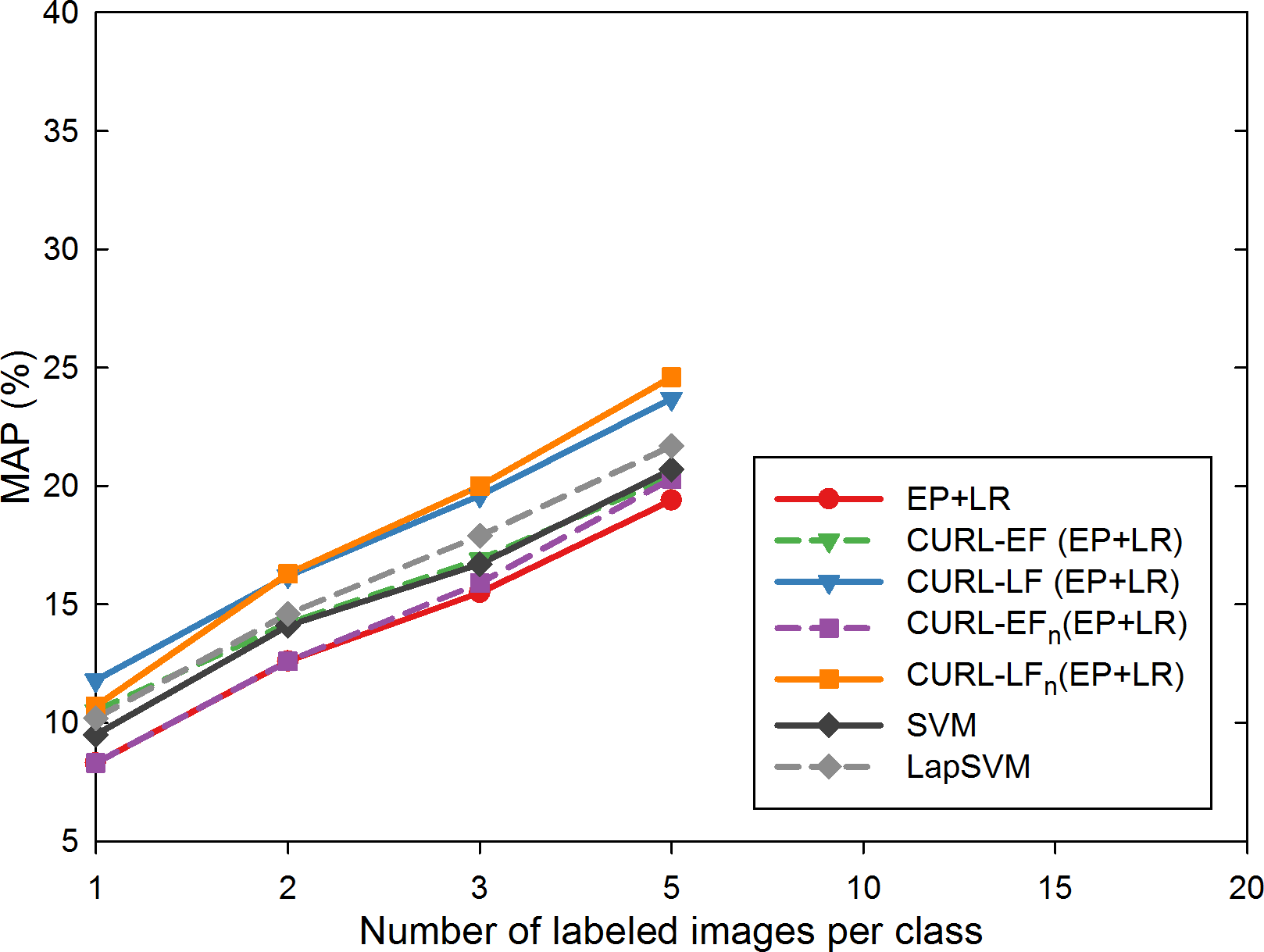} &
\includegraphics[width=0.68\columnwidth]{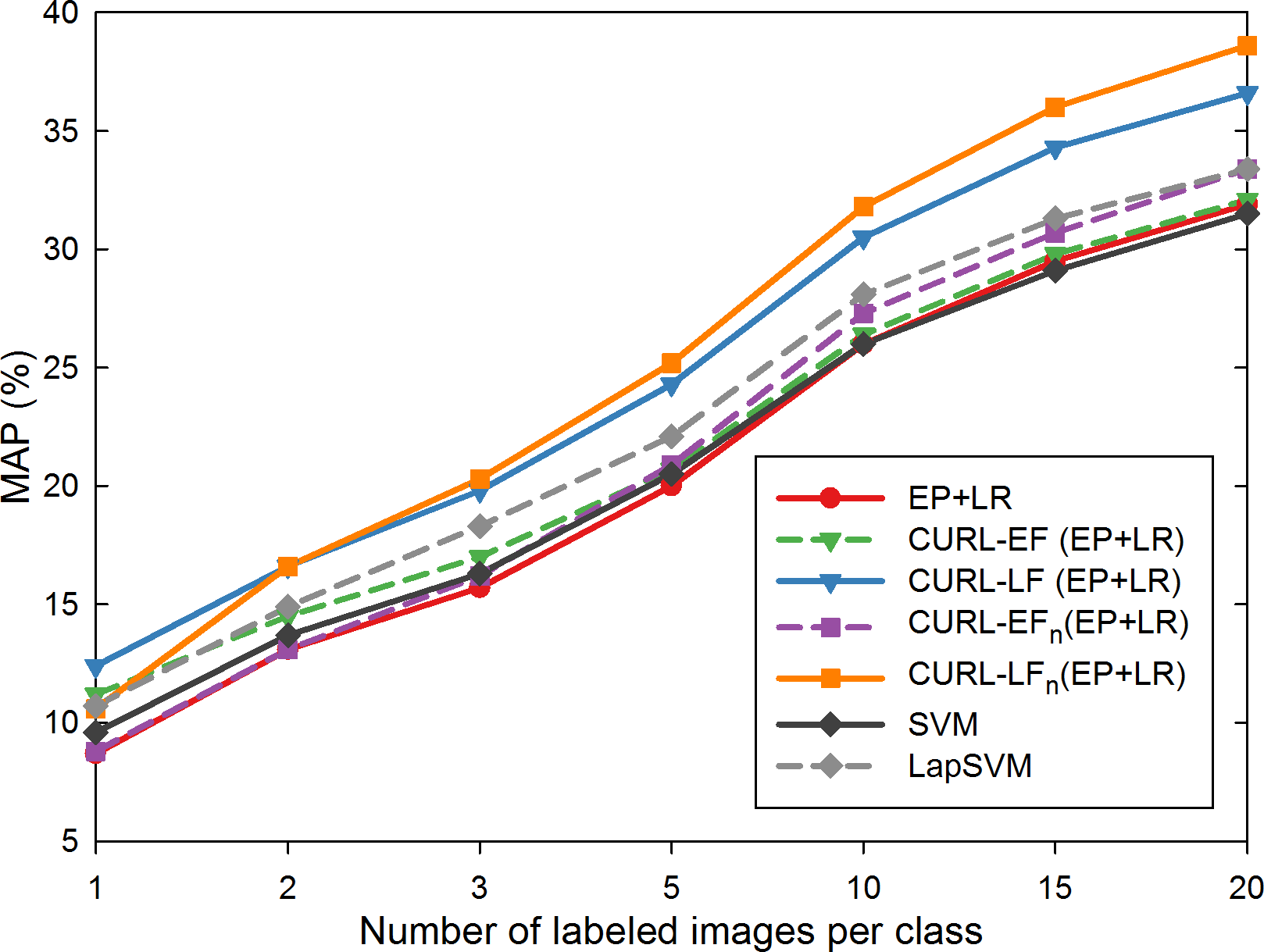} &
\includegraphics[width=0.68\columnwidth]{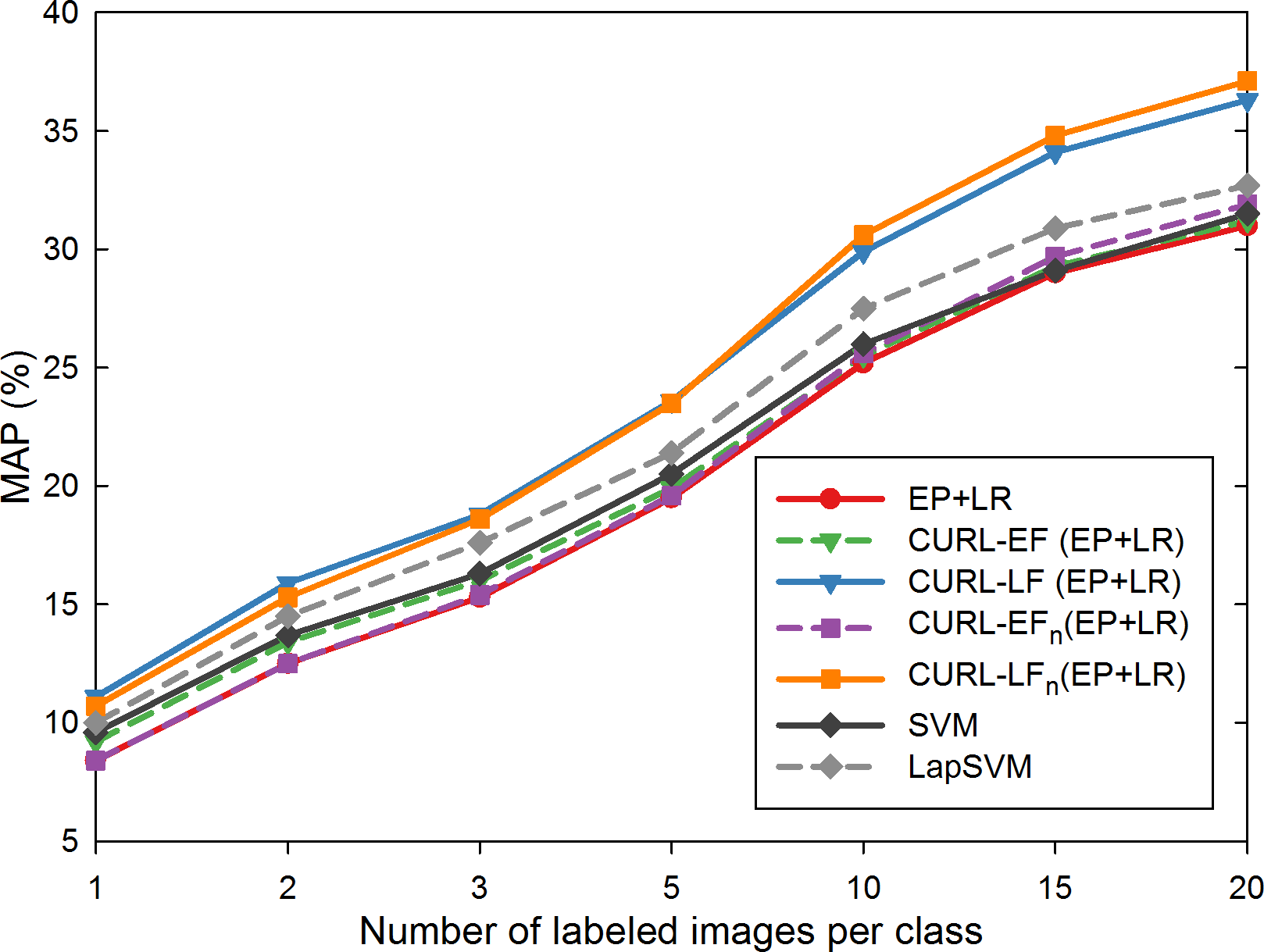} \\
\\
 
\end{tabular}
\caption{Mean Average Precision (MAP) varying the number of labeled
  images per class, obtained on the Scene-15 data set (first row), and
  on the Caltech-101 data set (second row).  Three scenarios are
  considered: inductive learning (left column), transductive learning
  (middle column) and self-taught learning (third column). Note that
  inductive learning on the Caltech-101 data set is limited to 5
  labeled images per class because otherwise for some classes there
  wouldn't be enough unlabeled data left for both training and
  evaluation.}
\label{fig:risultati}
\end{figure*}

Fig.~\ref{fig:risultati} shows the classification performance with
different numbers of labeled training images per class, in the three
learning scenarios for both the Scene-15 and Caltech-101
data sets. For the CURL-based methods we considered five co-training rounds, and the reported performance correspond to the last round. For SVM and LapSVM only the results
using $\chi^2$ kernel are reported, since they consistently showed 
the best performance across all the experiments. 
Detailed results for all the tested baseline methods, and for the CURL variants across the co-training rounds are available in Tables \ref{tab:SVMeLapSVM}, \ref{tab:exp15scenes} and \ref{tab:exp101caltech}. 
\begin{table}[!htbp]
\caption{Mean Average Precision (MAP) of the baseline algorithms, varying the number of labeled images per class in the three learning scenarios considered: inductive (IND), transductive (TRD), and self-taught (ST).
}
\label{tab:SVMeLapSVM}
\begin{center}
\resizebox{1.00\columnwidth}{!} {
\begin{tabular}{ccc}
\begin{tabular}{|r|l|rrr|}
    \hline
& & \multicolumn{3}{ c |}{Scene-15} \\ 
\# img & method & IND & TRD & ST \\
\hline
 \multirow{6}{*}{1}
& SVM$_{lin}$ 	 		& 22.7 & 22.3 & 22.3 \\
& SVM$_{rbf}$ 	 		& 26.5 & 25.8 & 25.8 \\
& SVM$_{\chi^2}$ 		& 29.5 & 28.7 & 28.7 \\
& LapSVM$_{lin}$ 	 	& 28.1 & 29.1 & 26.4 \\
& LapSVM$_{rbf}$ 	 	& 28.8 & 29.8 & 26.7 \\
& LapSVM$_{\chi^2}$ & 32.3 & 33.7 & 30.2 \\
& EP+LR							& 38.3 & 39.3 & 32.4 \\
\hline
\multirow{6}{*}{2}
& SVM$_{lin}$ 	 		& 27.4 & 26.9 & 26.9 \\
& SVM$_{rbf}$ 	 		& 32.9 & 31.3 & 31.3 \\
& SVM$_{\chi^2}$ 		& 35.4 & 34.9 & 34.9 \\
& LapSVM$_{lin}$ 	 	& 33.7 & 34.9 & 31.2 \\
& LapSVM$_{rbf}$ 	 	& 34.6 & 35.7 & 32.5 \\
& LapSVM$_{\chi^2}$ & 38.1 & 39.6 & 35.7 \\
& EP+LR							& 44.6 & 47.3 & 41.0 \\
\hline
\multirow{6}{*}{3}
& SVM$_{lin}$ 	 		& 30.0 & 30.2 & 30.2 \\
& SVM$_{rbf}$ 	 		& 36.5 & 36.7 & 36.7 \\
& SVM$_{\chi^2}$ 		& 39.9 & 39.3 & 39.3 \\
& LapSVM$_{lin}$ 	 	& 37.6 & 38.6 & 36.4 \\
& LapSVM$_{rbf}$ 	 	& 37.7 & 38.9 & 37.2 \\
& LapSVM$_{\chi^2}$ & 42.8 & 43.9 & 41.6 \\
& EP+LR							& 50.8 & 53.2 & 48.5 \\
\hline
\multirow{6}{*}{5}
& SVM$_{lin}$ 	 		& 35.5 & 35.2 & 35.2 \\
& SVM$_{rbf}$ 	 		& 43.5 & 42.8 & 42.8 \\
& SVM$_{\chi^2}$ 		& 46.5 & 46.1 & 46.1 \\
& LapSVM$_{lin}$ 	 	& 43.4 & 44.5 & 43.5 \\
& LapSVM$_{rbf}$ 	 	& 43.8 & 44.7 & 44.0 \\
& LapSVM$_{\chi^2}$ & 49.2 & 50.5 & 49.3 \\
& EP+LR							& 55.6 & 58.6 & 55.2 \\
\hline
\multirow{6}{*}{10}
& SVM$_{lin}$ 	 		& 43.8 & 43.7 & 43.7 \\
& SVM$_{rbf}$ 	 		& 51.9 & 51.3 & 51.3 \\
& SVM$_{\chi^2}$ 		& 55.5 & 55.2 & 55.2 \\
& LapSVM$_{lin}$ 	 	& 51.4 & 52.5 & 52.3 \\
& LapSVM$_{rbf}$ 	 	& 52.4 & 53.1 & 53.0 \\ 
& LapSVM$_{\chi^2}$ & 56.6 & 57.4 & 57.2 \\
& EP+LR							& 62.8 & 64.4 & 62.9 \\
\hline
\multirow{6}{*}{15}
& SVM$_{lin}$ 	 		& 50.1 & 50.3 & 50.3 \\
& SVM$_{rbf}$ 	 		& 57.4 & 57.1 & 57.1 \\
& SVM$_{\chi^2}$ 		& 60.7 & 60.6 & 60.6 \\
& LapSVM$_{lin}$ 	 	& 55.1 & 56.0 & 55.5 \\
& LapSVM$_{rbf}$ 	 	& 57.8 & 58.7 & 58.3 \\ 
& LapSVM$_{\chi^2}$ & 60.9 & 61.6 & 61.2 \\
& EP+LR							& 66.0 & 67.9 & 67.3 \\
\hline
\multirow{6}{*}{20}
& SVM$_{lin}$ 	 		& 54.0 & 53.5 & 53.5 \\
& SVM$_{rbf}$ 	 		& 60.4 & 60.3 & 60.3 \\
& SVM$_{\chi^2}$ 		& 64.2 & 64.1 & 64.1 \\
& LapSVM$_{lin}$ 	 	& 58.3 & 58.7 & 58.6 \\
& LapSVM$_{rbf}$ 	 	& 60.8 & 61.4 & 61.2 \\ 
& LapSVM$_{\chi^2}$ & 64.5 & 65.4 & 65.1 \\ 
& EP+LR							& 67.8 & 69.7 & 69.1 \\
\hline
\end{tabular}

&

\begin{tabular}{|r|l|rrr|}
    \hline
& & \multicolumn{3}{ c |}{Caltech-101} \\ 
\# img & method & IND & TRD & ST \\
\hline
 \multirow{6}{*}{1}
& SVM$_{lin}$ 	 		&  6.0 &  6.3 &  6.3 \\
& SVM$_{rbf}$ 	 		&  7.1 &  7.3 &  7.3 \\
& SVM$_{\chi^2}$ 		&  9.5 &  9.6 &  9.6 \\
& LapSVM$_{lin}$ 	 	&  9.2 &  9.6 &  9.1 \\
& LapSVM$_{rbf}$ 	 	&  9.8 & 10.2 &  9.6 \\
& LapSVM$_{\chi^2}$ & 10.2 & 10.7 & 10.0 \\ 
& EP+LR							&  8.3 &  8.7 &  8.4 \\
\hline
\multirow{6}{*}{2}
& SVM$_{lin}$ 	 		&  9.1 &  9.2 &  9.2 \\
& SVM$_{rbf}$ 	 		& 10.1 &  9.8 &  9.8 \\
& SVM$_{\chi^2}$ 		& 14.1 & 13.7 & 13.7 \\
& LapSVM$_{lin}$ 	 	& 12.3 & 12.8 & 12.1 \\
& LapSVM$_{rbf}$ 	 	& 12.7 & 13.3 & 12.4 \\
& LapSVM$_{\chi^2}$ & 14.6 & 14.9 & 14.5 \\ 
& EP+LR							& 12.6 & 13.1 & 12.5 \\
\hline
\multirow{6}{*}{3}
& SVM$_{lin}$ 	 		& 10.7 & 10.8 & 10.8 \\
& SVM$_{rbf}$ 	 		& 11.7 & 11.6 & 11.6 \\
& SVM$_{\chi^2}$ 		& 16.7 & 16.3 & 16.3 \\
& LapSVM$_{lin}$ 	 	& 13.8 & 14.3 & 13.5 \\
& LapSVM$_{rbf}$ 	 	& 14.0 & 14.6 & 13.9 \\
& LapSVM$_{\chi^2}$ & 17.9 & 18.3 & 17.6 \\
& EP+LR							& 15.5 & 15.7 & 15.3 \\
\hline
\multirow{6}{*}{5}
& SVM$_{lin}$ 	 		& 13.4 & 13.3 & 13.3 \\
& SVM$_{rbf}$ 	 		& 14.9 & 14.6 & 14.6 \\
& SVM$_{\chi^2}$ 		& 20.7 & 20.5 & 20.5 \\
& LapSVM$_{lin}$ 	 	& 16.3 & 16.6 & 16.0 \\
& LapSVM$_{rbf}$ 	 	& 16.8 & 17.1 & 16.6 \\
& LapSVM$_{\chi^2}$ & 21.7 & 22.1 & 21.4 \\
& EP+LR							& 19.4 & 20.0 & 19.5 \\
\hline
\multirow{6}{*}{10}
& SVM$_{lin}$ 	 		& - & 17.3 & 17.3 \\
& SVM$_{rbf}$ 	 		& - & 19.2 & 19.2 \\
& SVM$_{\chi^2}$ 		& - & 26.0 & 26.0 \\
& LapSVM$_{lin}$ 	 	& - & 20.5 & 20.1 \\
& LapSVM$_{rbf}$ 	 	& - & 21.6 & 21.2 \\
& LapSVM$_{\chi^2}$ & - & 28.1 & 27.5 \\
& EP+LR							& - & 26.0 & 25.2 \\
\hline
\multirow{6}{*}{15}
& SVM$_{lin}$ 	 		& - & 19.7 & 19.7 \\
& SVM$_{rbf}$ 	 		& - & 22.1 & 22.1 \\
& SVM$_{\chi^2}$ 		& - & 29.1 & 29.1 \\
& LapSVM$_{lin}$ 	 	& - & 22.9 & 22.5 \\
& LapSVM$_{rbf}$ 	 	& - & 24.0 & 23.5 \\
& LapSVM$_{\chi^2}$ & - & 31.3 & 30.9 \\
& EP+LR							& - & 29.5 & 29.0 \\
\hline
\multirow{6}{*}{20}
& SVM$_{lin}$ 	 		& - & 21.5 & 21.5 \\
& SVM$_{rbf}$ 	 		& - & 24.2 & 24.2 \\
& SVM$_{\chi^2}$ 		& - & 31.5 & 31.5 \\
& LapSVM$_{lin}$ 	 	& - & 24.6 & 24.0 \\
& LapSVM$_{rbf}$ 	 	& - & 26.8 & 26.1 \\
& LapSVM$_{\chi^2}$ & - & 33.4 & 32.7 \\
& EP+LR							& - & 31.9 & 31.0 \\
\hline
\end{tabular}

\\
\end{tabular}

}
\end{center}
\end{table}

\begin{table*}[!ht]
\caption{Mean Average Precision (MAP) \ADD{ of the CURL variants, in the (EP+LR) \emb,} varying the number of labeled images per class at the different co-training rounds obtained on the Scene-15 data set in the three learning scenarios considered: inductive (left), transductive (middle), and self-taught (right). \ADD{For clarity, the (EP+LR) suffixes have been omitted.}}
\label{tab:exp15scenes}
\begin{center}
\resizebox{\linewidth}{!} {
\begin{tabular}{ccc}

\begin{tabular}{|r|l|rrrrrr|}
    \hline
& & \multicolumn{6}{ c |}{\# co-train round} \\ 
\# img & method & 0 & 1 & 2 & 3 & 4 & 5 \\
\hline
    \multirow{6}{*}{1}
& CURL-EF 					& 38.3 & 41.0 & 41.1 & 41.1 & 41.1 & 41.2 \\
& CURL-LF 					& 40.0 & 42.4 & 43.8 & 44.0 & 44.3 & 44.4 \\
& CURL-EF\&LF 	& - & 43.3 & 43.7 & 43.8 & 44.0 & 44.0 \\
& CURL-EF$_n$ 					& 38.3 & 38.4 & 38.5 & 38.4 & 38.3 & 38.2 \\
& CURL-LF$_n$ 					& 40.0 & 39.9 & 39.9 & 39.7 & 39.6 & 39.4 \\
& CURL-EF\&LF$_n$ 	&- & 40.4 & 40.4 & 40.3 & 40.1 & 40.0 \\
\hline
\multirow{6}{*}{2}
& CURL-EF & 44.6 & 46.3 & 46.4 & 46.5 & 46.6 & 46.6 \\
& CURL-LF& 48.5 & 50.2 & 50.7 & 50.9 & 51.1 & 51.3 \\
& CURL-EF\&LF& - & 49.7 & 50.0 & 50.1 & 50.2 & 50.3 \\
& CURL-EF$_n$& 44.6 & 44.3 & 44.7 & 44.7 & 44.5 & 44.2 \\
& CURL-LF$_n$& 48.5 & 48.8 & 48.8 & 48.4 & 48.0 & 47.3 \\
& CURL-EF\&LF$_n$& - & 48.0 & 48.0 & 47.8 & 47.4 & 46.9 \\
\hline
\multirow{6}{*}{3}
& CURL-EF & 50.8 & 52.1 & 52.2 & 52.3 & 52.4 & 52.4 \\
& CURL-LF& 54.1 & 55.2 & 56.2 & 56.4 & 56.6 & 56.8 \\
& CURL-EF\&LF& - & 55.3 & 55.7 & 55.9 & 56.0 & 56.1 \\
& CURL-EF$_n$& 50.8 & 51.6 & 52.5 & 52.6 & 52.0 & 51.8 \\
& CURL-LF$_n$& 54.1 & 55.3 & 55.3 & 55.1 & 54.7 & 54.2 \\
& CURL-EF\&LF$_n$& - & 55.0 & 55.2 & 55.0 & 54.4 & 54.0 \\
\hline
\multirow{6}{*}{5}
& CURL-EF & 55.6 & 56.1 & 56.2 & 56.2 & 56.2 & 56.2 \\
& CURL-LF& 58.3 & 59.0 & 59.3 & 59.5 & 59.5 & 59.6 \\
& CURL-EF\&LF& - & 59.0 & 59.1 & 59.2 & 59.3 & 59.4 \\
& CURL-EF$_n$& 55.6 & 56.5 & 57.7 & 57.7 & 57.7 & 57.6 \\
& CURL-LF$_n$& 58.3 & 59.7 & 59.8 & 60.0 & 59.8 & 59.8 \\
& CURL-EF\&LF$_n$& - & 59.6 & 59.9 & 59.9 & 59.7 & 59.6 \\
\hline
\multirow{6}{*}{10}
& CURL-EF & 62.8 & 63.1 & 63.2 & 63.2 & 63.2 & 63.2 \\
& CURL-LF& 66.2 & 66.5 & 66.6 & 66.9 & 67.0 & 67.1 \\
& CURL-EF\&LF& - & 66.3 & 66.4 & 66.5 & 66.6 & 66.7 \\
& CURL-EF$_n$& 62.8 & 64.7 & 65.3 & 65.5 & 65.8 & 65.8 \\
& CURL-LF$_n$& 66.2 & 67.6 & 67.8 & 68.2 & 68.2 & 68.3 \\
& CURL-EF\&LF$_n$& - & 67.5 & 67.8 & 68.0 & 68.0 & 68.0 \\
\hline
\multirow{6}{*}{15}
& CURL-EF & 66.0 & 66.1 & 66.1 & 66.1 & 66.1 & 66.1 \\
& CURL-LF& 69.2 & 69.4 & 69.6 & 69.8 & 69.8 & 69.9 \\
& CURL-EF\&LF& - & 69.3 & 69.3 & 69.4 & 69.4 & 69.5 \\
& CURL-EF$_n$& 66.0 & 67.6 & 68.3 & 68.5 & 68.7 & 68.8 \\
& CURL-LF$_n$& 69.2 & 70.6 & 70.9 & 71.0 & 71.0 & 70.9 \\
& CURL-EF\&LF$_n$& - & 70.4 & 70.8 & 70.8 & 70.9 & 70.9 \\
\hline
\multirow{6}{*}{20}
& CURL-EF & 67.8 & 67.9 & 67.9 & 67.9 & 67.9 & 67.9 \\
& CURL-LF& 71.1 & 71.3 & 71.3 & 71.5 & 71.5 & 71.6 \\
& CURL-EF\&LF& - & 71.2 & 71.2 & 71.2 & 71.2 & 71.3 \\
& CURL-EF$_n$& 67.8 & 69.2 & 69.8 & 70.1 & 70.2 & 70.3 \\
& CURL-LF$_n$& 71.1 & 72.0 & 72.4 & 72.5 & 72.5 & 72.5 \\
& CURL-EF\&LF$_n$& - & 71.9 & 72.2 & 72.4 & 72.4 & 72.4 \\
		\hline
\end{tabular}

&

\begin{tabular}{|r|l|rrrrrr|}
    \hline
& & \multicolumn{6}{ c |}{\# co-train round} \\ 
\# img & method & 0 & 1 & 2 & 3 & 4 & 5 \\
\hline
\multirow{6}{*}{1}
& CURL-EF & 39.3 & 41.7 & 41.9 & 41.9 & 41.9 & 41.9 \\
& CURL-LF & 39.8 & 42.6 & 43.8 & 44.1 & 44.2 & 44.4 \\
& CURL-EF\&LF & - & 43.5 & 43.9 & 44.1 & 44.1 & 44.2 \\
& CURL-EF$_n$ & 39.3 & 38.9 & 38.9 & 38.9 & 38.7 & 38.7 \\
& CURL-LF$_n$ & 39.8 & 39.9 & 39.9 & 39.7 & 39.6 & 39.3 \\
& CURL-EF\&LF$_n$ & - & 40.4 & 40.4 & 40.4 & 40.2 & 40.0 \\
\hline
\multirow{6}{*}{2}
& CURL-EF & 47.3 & 49.2 & 49.5 & 49.6 & 49.7 & 49.7 \\
& CURL-LF & 48.7 & 50.6 & 51.5 & 51.9 & 52.0 & 52.1 \\
& CURL-EF\&LF & - & 51.2 & 51.7 & 51.9 & 52.0 & 52.1 \\
& CURL-EF$_n$ & 47.3 & 47.4 & 47.8 & 47.5 & 47.1 & 46.9 \\
& CURL-LF$_n$ & 48.7 & 49.3 & 49.0 & 48.8 & 48.5 & 47.8 \\
& CURL-EF\&LF$_n$ & - & 49.6 & 49.5 & 49.2 & 48.8 & 48.2 \\
\hline
\multirow{6}{*}{3}
& CURL-EF & 53.2 & 54.4 & 54.5 & 54.6 & 54.7 & 54.7 \\
& CURL-LF & 54.8 & 55.8 & 56.6 & 57.2 & 57.4 & 57.5 \\
& CURL-EF\&LF & - & 56.4 & 56.8 & 57.2 & 57.3 & 57.4 \\
& CURL-EF$_n$ & 53.2 & 53.4 & 53.7 & 53.5 & 52.7 & 52.2 \\
& CURL-LF$_n$ & 54.8 & 55.4 & 55.4 & 54.9 & 54.4 & 53.7 \\
& CURL-EF\&LF$_n$ & - & 55.8 & 55.7 & 55.2 & 54.5 & 53.9 \\
\hline
\multirow{6}{*}{5}
& CURL-EF & 58.6 & 59.1 & 59.2 & 59.2 & 59.2 & 59.2 \\
& CURL-LF & 60.3 & 61.3 & 61.6 & 61.7 & 61.9 & 61.9 \\
& CURL-EF\&LF & - & 61.5 & 61.6 & 61.7 & 61.8 & 61.8 \\
& CURL-EF$_n$ & 58.6 & 59.6 & 60.4 & 60.3 & 60.0 & 59.7 \\
& CURL-LF$_n$ & 60.3 & 61.8 & 62.0 & 61.9 & 61.7 & 61.5 \\
& CURL-EF\&LF$_n$ & - & 61.9 & 62.1 & 61.9 & 61.6 & 61.3 \\
\hline
\multirow{6}{*}{10}
& CURL-EF & 64.4 & 64.6 & 64.6 & 64.6 & 64.7 & 64.7 \\
& CURL-LF & 66.6 & 67.0 & 67.2 & 67.3 & 67.5 & 67.6 \\
& CURL-EF\&LF & - & 67.1 & 67.1 & 67.2 & 67.3 & 67.4 \\
& CURL-EF$_n$ & 64.4 & 66.2 & 67.0 & 67.3 & 67.4 & 67.5 \\
& CURL-LF$_n$ & 66.6 & 68.5 & 68.8 & 69.0 & 69.0 & 69.0 \\
& CURL-EF\&LF$_n$ & - & 68.4 & 68.8 & 69.0 & 69.0 & 69.0 \\
\hline
\multirow{6}{*}{15}
& CURL-EF & 67.9 & 68.0 & 68.0 & 68.0 & 68.0 & 68.0 \\
& CURL-LF & 70.1 & 70.5 & 70.6 & 70.7 & 70.8 & 71.0 \\
& CURL-EF\&LF & - & 70.5 & 70.6 & 70.6 & 70.7 & 70.8 \\
& CURL-EF$_n$ & 67.9 & 69.5 & 70.4 & 70.7 & 70.8 & 70.9 \\
& CURL-LF$_n$ & 70.1 & 71.7 & 71.9 & 72.2 & 72.3 & 72.3 \\
& CURL-EF\&LF$_n$ & - & 71.7 & 72.1 & 72.3 & 72.4 & 72.4 \\
\hline
\multirow{6}{*}{20}
& CURL-EF & 69.7 & 69.8 & 69.8 & 69.8 & 69.8 & 69.8 \\
& CURL-LF & 72.1 & 72.2 & 72.3 & 72.4 & 72.5 & 72.6 \\
& CURL-EF\&LF & - & 72.3 & 72.3 & 72.4 & 72.4 & 72.4 \\
& CURL-EF$_n$ & 69.7 & 71.2 & 71.7 & 72.1 & 72.2 & 72.3 \\
& CURL-LF$_n$ & 72.1 & 73.3 & 73.6 & 73.7 & 73.7 & 73.7 \\
& CURL-EF\&LF$_n$ & - & 73.3 & 73.6 & 73.7 & 73.8 & 73.8 \\
\hline
\end{tabular}

& 

\begin{tabular}{|r|l|rrrrrr|}
    \hline
& & \multicolumn{6}{ c |}{\# co-train round} \\ 
\# img & method & 0 & 1 & 2 & 3 & 4 & 5 \\
\hline
\multirow{6}{*}{1}
& CURL-EF & 32.4 & 34.8 & 35.2 & 35.3 & 35.3 & 35.3 \\
& CURL-LF & 36.7 & 38.6 & 39.6 & 39.7 & 39.8 & 39.9 \\
& CURL-EF\&LF & - & 38.4 & 38.9 & 38.9 & 39.0 & 39.0 \\
& CURL-EF$_n$ & 32.4 & 32.4 & 32.8 & 32.8 & 32.8 & 32.3 \\
& CURL-LF$_n$ & 36.7 & 34.9 & 34.8 & 34.8 & 34.6 & 34.2 \\
& CURL-EF\&LF$_n$ & - & 35.5 & 35.5 & 35.6 & 35.4 & 35.0 \\
\hline
\multirow{6}{*}{2}
& CURL-EF & 41.0 & 42.4 & 42.5 & 42.6 & 42.6 & 42.7 \\
& CURL-LF & 45.4 & 47.1 & 47.6 & 47.5 & 47.7 & 47.7 \\
& CURL-EF\&LF & - & 46.4 & 46.5 & 46.6 & 46.7 & 46.7 \\
& CURL-EF$_n$ & 41.0 & 41.2 & 42.7 & 42.5 & 41.6 & 41.4 \\
& CURL-LF$_n$ & 45.4 & 44.3 & 44.2 & 43.7 & 43.1 & 42.3 \\
& CURL-EF\&LF$_n$ & - & 44.8 & 45.1 & 44.7 & 43.7 & 43.0 \\
\hline
\multirow{6}{*}{3}
& CURL-EF & 48.5 & 49.4 & 49.5 & 49.5 & 49.5 & 49.6 \\
& CURL-LF & 53.4 & 54.0 & 54.5 & 54.4 & 54.5 & 54.5 \\
& CURL-EF\&LF & - & 53.5 & 53.6 & 53.6 & 53.6 & 53.6 \\
& CURL-EF$_n$ & 48.5 & 49.9 & 50.6 & 50.6 & 49.8 & 49.3 \\
& CURL-LF$_n$ & 53.4 & 52.0 & 52.3 & 52.0 & 51.6 & 51.1 \\
& CURL-EF\&LF$_n$ & - & 52.8 & 52.9 & 52.6 & 51.9 & 51.3 \\
\hline
\multirow{6}{*}{5}
& CURL-EF & 55.2 & 55.4 & 55.5 & 55.5 & 55.5 & 55.4 \\
& CURL-LF & 59.4 & 59.8 & 60.0 & 60.1 & 60.1 & 60.1 \\
& CURL-EF\&LF & - & 59.3 & 59.4 & 59.4 & 59.4 & 59.4 \\
& CURL-EF$_n$ & 55.2 & 56.6 & 57.5 & 57.5 & 57.2 & 57.1 \\
& CURL-LF$_n$ & 59.4 & 59.5 & 59.6 & 59.6 & 59.4 & 59.2 \\
& CURL-EF\&LF$_n$ & - & 59.8 & 59.8 & 59.7 & 59.3 & 59.1 \\
\hline
\multirow{6}{*}{10}
& CURL-EF & 62.9 & 63.1 & 63.1 & 63.1 & 63.1 & 63.1 \\
& CURL-LF & 66.0 & 66.4 & 66.7 & 66.7 & 66.9 & 67.0 \\
& CURL-EF\&LF & - & 66.6 & 66.7 & 66.7 & 66.7 & 66.8 \\
& CURL-EF$_n$ & 62.9 & 64.7 & 65.7 & 66.0 & 66.1 & 66.3 \\
& CURL-LF$_n$ & 66.0 & 67.1 & 67.3 & 67.5 & 67.6 & 67.6 \\
& CURL-EF\&LF$_n$ & - & 67.2 & 67.6 & 67.7 & 67.8 & 67.8 \\
\hline
\multirow{6}{*}{15}
& CURL-EF & 67.3 & 67.5 & 67.5 & 67.5 & 67.5 & 67.5 \\
& CURL-LF & 70.5 & 70.8 & 70.9 & 71.1 & 71.2 & 71.3 \\
& CURL-EF\&LF & - & 71.0 & 71.0 & 71.1 & 71.2 & 71.2 \\
& CURL-EF$_n$ & 67.3 & 69.2 & 69.8 & 70.1 & 70.2 & 70.2 \\
& CURL-LF$_n$ & 70.5 & 70.8 & 71.1 & 71.3 & 71.5 & 71.5 \\
& CURL-EF\&LF$_n$ & - & 71.3 & 71.5 & 71.7 & 71.8 & 71.8 \\
\hline
\multirow{6}{*}{20}
& CURL-EF & 69.1 & 69.2 & 69.2 & 69.2 & 69.2 & 69.2 \\
& CURL-LF & 72.3 & 72.4 & 72.4 & 72.5 & 72.6 & 72.8 \\
& CURL-EF\&LF & - & 72.6 & 72.6 & 72.7 & 72.7 & 72.8 \\
& CURL-EF$_n$ & 69.1 & 70.8 & 71.5 & 71.7 & 71.8 & 71.9 \\
& CURL-LF$_n$ & 72.3 & 72.4 & 72.7 & 72.8 & 72.8 & 72.8 \\
& CURL-EF\&LF$_n$ & - & 72.9 & 73.2 & 73.2 & 73.3 & 73.3 \\
\hline
\end{tabular}

\\
\end{tabular}

}
\end{center}
\end{table*}

\begin{table*}[!ht]
\caption{Mean Average Precision (MAP) \ADD{of the CURL variants, in the (EP+LR) \emb,} varying the number of labeled images per class at the different co-training rounds obtained on the Caltech-101 data set in the three learning scenarios considered: inductive (left), transductive  (middle), and self-taught (right). \ADD{For clarity, the (EP+LR) suffixes have been omitted.}}
\label{tab:exp101caltech}
\begin{center}
\resizebox{\linewidth}{!} {

\begin{tabular}{ccc}

\begin{tabular}{|r|l|rrrrrr|}
    \hline
& & \multicolumn{6}{ c |}{\# co-train round} \\ 
\# img & method & 0 & 1 & 2 & 3 & 4 & 5 \\
\hline
 \multirow{6}{*}{1}
& CURL-EF & 8.3 & 10.4 & 10.5 & 10.5 & 10.5 & 10.5 \\
& CURL-LF & 10.1 & 11.6 & 11.8 & 11.8 & 11.8 & 11.8 \\
& CURL-EF\&LF & - & 11.5 & 11.6 & 11.6 & 11.6 & 11.6 \\
& CURL-EF$_n$ & 8.3 & 8.3 & 8.5 & 8.5 & 8.8 & 8.8 \\
& CURL-LF$_n$ & 10.1 & 10.3 & 10.3 & 10.6 & 10.7 & 10.7 \\
& CURL-EF\&LF$_n$ & - & 9.5 & 9.6 & 9.8 & 10.0 & 10.1 \\
\hline
\multirow{6}{*}{2}
& CURL-EF & 12.6 & 14.2 & 14.3 & 14.3 & 14.3 & 14.2 \\
& CURL-LF & 15.3 & 16.3 & 16.3 & 16.3 & 16.2 & 16.2 \\
& CURL-EF\&LF & - & 15.9 & 15.9 & 15.8 & 15.8 & 15.8 \\
& CURL-EF$_n$ & 12.6 & 12.6 & 13.4 & 13.5 & 13.9 & 14.1 \\
& CURL-LF$_n$ & 15.3 & 15.7 & 15.9 & 16.1 & 16.2 & 16.3 \\
& CURL-EF\&LF$_n$ & - & 14.6 & 15.1 & 15.3 & 15.5 & 15.7 \\
\hline
\multirow{6}{*}{3}
& CURL-EF & 15.5 & 16.8 & 16.9 & 16.8 & 16.9 & 16.9 \\
& CURL-LF & 18.8 & 19.4 & 19.6 & 19.6 & 19.7 & 19.6 \\
& CURL-EF\&LF & - & 18.9 & 19.0 & 18.9 & 18.9 & 18.9 \\
& CURL-EF$_n$ & 15.5 & 15.9 & 16.8 & 16.9 & 17.1 & 17.1 \\
& CURL-LF$_n$ & 18.8 & 19.5 & 19.8 & 20.0 & 20.0 & 20.0 \\
& CURL-EF\&LF$_n$ & - & 18.5 & 19.0 & 19.2 & 19.3 & 19.2 \\
\hline
\multirow{6}{*}{5}
& CURL-EF & 19.4 & 20.2 & 20.4 & 20.4 & 20.4 & 20.4 \\
& CURL-LF & 23.2 & 23.4 & 23.7 & 23.7 & 23.7 & 23.7 \\
& CURL-EF\&LF & - & 22.9 & 23.0 & 23.0 & 23.0 & 23.0 \\
& CURL-EF$_n$ & 19.4 & 20.3 & 21.0 & 21.2 & 21.3 & 21.3 \\
& CURL-LF$_n$ & 23.2 & 24.2 & 24.4 & 24.6 & 24.6 & 24.6 \\
& CURL-EF\&LF$_n$ & - & 23.3 & 23.7 & 23.9 & 23.9 & 24.0 \\
\hline
\multirow{6}{*}{10}
& CURL-EF & - & - & - & - & - & - \\
& CURL-LF & - & - & - & - & - & - \\
& CURL-EF\&LF & - & - & - & - & - & - \\
& CURL-EF$_n$ & - & - & - & - & - & - \\
& CURL-LF$_n$ & - & - & - & - & - & - \\
& CURL-EF\&LF$_n$ & - & - & - & - & - & - \\
\hline
\multirow{6}{*}{15}
& CURL-EF & - & - & - & - & - & - \\
& CURL-LF & - & - & - & - & - & - \\
& CURL-EF\&LF & - & - & - & - & - & - \\
& CURL-EF$_n$ & - & - & - & - & - & - \\
& CURL-LF$_n$ & - & - & - & - & - & - \\
& CURL-EF\&LF$_n$ & - & - & - & - & - & - \\
\hline
\multirow{6}{*}{20}
& CURL-EF & - & - & - & - & - & - \\
& CURL-LF & - & - & - & - & - & - \\
& CURL-EF\&LF & - & - & - & - & - & - \\
& CURL-EF$_n$ & - & - & - & - & - & - \\
& CURL-LF$_n$ & - & - & - & - & - & - \\
& CURL-EF\&LF$_n$ & - & - & - & - & - & - \\
\hline
\end{tabular}

&

\begin{tabular}{|r|l|rrrrrr|}
    \hline
& & \multicolumn{6}{ c |}{\# co-train round} \\ 
\# img & method & 0 & 1 & 2 & 3 & 4 & 5 \\
\hline
 \multirow{6}{*}{1}
& CURL-EF & 8.7 & 11.1 & 11.2 & 11.2 & 11.2 & 11.2 \\
& CURL-LF & 10.5 & 12.2 & 12.4 & 12.4 & 12.4 & 12.4 \\
& CURL-EF\&LF & - & 12.1 & 12.2 & 12.2 & 12.2 & 12.2 \\
& CURL-EF$_n$ & 8.7 & 8.8 & 8.8 & 8.8 & 8.9 & 8.9 \\
& CURL-LF$_n$ & 10.5 & 10.5 & 10.5 & 10.6 & 10.6 & 10.6 \\
& CURL-EF\&LF$_n$ & - & 9.8 & 9.8 & 9.9 & 9.9 & 9.9 \\
\hline
\multirow{6}{*}{2}
& CURL-EF & 13.1 & 14.4 & 14.5 & 14.6 & 14.5 & 14.5 \\
& CURL-LF & 15.7 & 16.5 & 16.6 & 16.6 & 16.6 & 16.6 \\
& CURL-EF\&LF & - & 16.0 & 16.1 & 16.1 & 16.1 & 16.0 \\
& CURL-EF$_n$ & 13.1 & 13.1 & 13.8 & 13.9 & 14.1 & 14.2 \\
& CURL-LF$_n$ & 15.7 & 16.2 & 16.4 & 16.5 & 16.6 & 16.6 \\
& CURL-EF\&LF$_n$ & - & 15.2 & 15.5 & 15.7 & 15.8 & 15.8 \\
\hline
\multirow{6}{*}{3}
& CURL-EF & 15.7 & 16.8 & 16.9 & 16.9 & 17.0 & 17.0 \\
& CURL-LF & 19.0 & 19.5 & 19.8 & 19.8 & 19.8 & 19.8 \\
& CURL-EF\&LF & - & 18.9 & 19.0 & 19.0 & 19.0 & 19.0 \\
& CURL-EF$_n$ & 15.7 & 16.2 & 17.2 & 17.3 & 17.5 & 17.5 \\
& CURL-LF$_n$ & 19.0 & 20.0 & 20.2 & 20.3 & 20.3 & 20.3 \\
& CURL-EF\&LF$_n$ & - & 18.8 & 19.3 & 19.4 & 19.5 & 19.5 \\
\hline
\multirow{6}{*}{5}
& CURL-EF & 20.0 & 20.6 & 20.7 & 20.8 & 20.7 & 20.7 \\
& CURL-LF & 23.7 & 24.0 & 24.2 & 24.3 & 24.3 & 24.3 \\
& CURL-EF\&LF & - & 23.2 & 23.3 & 23.4 & 23.4 & 23.4 \\
& CURL-EF$_n$ & 20.0 & 20.9 & 21.8 & 21.9 & 22.0 & 22.0 \\
& CURL-LF$_n$ & 23.7 & 24.9 & 25.1 & 25.2 & 25.2 & 25.2 \\
& CURL-EF\&LF$_n$ & - & 23.8 & 24.3 & 24.4 & 24.4 & 24.4 \\
\hline
\multirow{6}{*}{10}
& CURL-EF & 26.0 & 26.3 & 26.4 & 26.4 & 26.4 & 26.4 \\
& CURL-LF & 30.2 & 30.2 & 30.4 & 30.4 & 30.5 & 30.5 \\
& CURL-EF\&LF & - & 29.6 & 29.6 & 29.6 & 29.7 & 29.7 \\
& CURL-EF$_n$ & 26.0 & 27.3 & 27.8 & 27.9 & 27.9 & 27.9 \\
& CURL-LF$_n$ & 30.2 & 31.6 & 31.8 & 31.8 & 31.8 & 31.8 \\
& CURL-EF\&LF$_n$ & - & 30.7 & 31.0 & 31.1 & 31.0 & 31.0 \\
\hline
\multirow{6}{*}{15}
& CURL-EF & 29.5 & 29.7 & 29.8 & 29.8 & 29.8 & 29.8 \\
& CURL-LF & 34.0 & 34.0 & 34.1 & 34.2 & 34.2 & 34.3 \\
& CURL-EF\&LF & - & 33.4 & 33.5 & 33.5 & 33.5 & 33.5 \\
& CURL-EF$_n$ & 29.5 & 30.7 & 31.1 & 31.3 & 31.3 & 31.3 \\
& CURL-LF$_n$ & 34.0 & 35.7 & 35.9 & 36.0 & 36.0 & 36.0 \\
& CURL-EF\&LF$_n$ & - & 34.7 & 35.0 & 35.1 & 35.1 & 35.1 \\
\hline
\multirow{6}{*}{20}
& CURL-EF & 31.9 & 32.1 & 32.1 & 32.1 & 32.1 & 32.1 \\
& CURL-LF & 36.4 & 36.4 & 36.5 & 36.6 & 36.6 & 36.6 \\
& CURL-EF\&LF & - & 36.1 & 36.1 & 36.1 & 36.0 & 36.0 \\
& CURL-EF$_n$ & 31.9 & 33.4 & 33.8 & 33.9 & 34.0 & 34.0 \\
& CURL-LF$_n$ & 36.4 & 38.1 & 38.4 & 38.5 & 38.6 & 38.6 \\
& CURL-EF\&LF$_n$ & - & 37.5 & 37.8 & 37.9 & 37.9 & 37.9 \\
\hline
\end{tabular}

&

\begin{tabular}{|r|l|rrrrrr|}
    \hline
& & \multicolumn{6}{ c |}{\# co-train round} \\ 
\# img & method & 0 & 1 & 2 & 3 & 4 & 5 \\
\hline
\multirow{6}{*}{1}
& CURL-EF & 8.4 & 9.3 & 9.3 & 9.2 & 9.2 & 9.2 \\
& CURL-LF & 10.7 & 11.4 & 11.3 & 11.2 & 11.1 & 11.1 \\
& CURL-EF\&LF & - & 10.9 & 10.8 & 10.7 & 10.7 & 10.7 \\
& CURL-EF$_n$ & 8.4 & 8.4 & 8.4 & 8.4 & 8.4 & 8.4 \\
& CURL-LF$_n$ & 10.7 & 10.7 & 10.7 & 10.7 & 10.7 & 10.7 \\
& CURL-EF\&LF$_n$ & - & 10.1 & 10.1 & 10.1 & 10.1 & 10.1 \\
\hline
\multirow{6}{*}{2}
& CURL-EF & 12.5 & 13.3 & 13.3 & 13.4 & 13.4 & 13.4 \\
& CURL-LF & 15.4 & 16.0 & 16.0 & 15.9 & 15.9 & 15.9 \\
& CURL-EF\&LF & - & 15.5 & 15.4 & 15.4 & 15.4 & 15.4 \\
& CURL-EF$_n$ & 12.5 & 12.5 & 12.5 & 12.5 & 12.5 & 12.5 \\
& CURL-LF$_n$ & 15.4 & 15.3 & 15.3 & 15.4 & 15.4 & 15.3 \\
& CURL-EF\&LF$_n$ & - & 14.7 & 14.7 & 14.7 & 14.7 & 14.7 \\
\hline
\multirow{6}{*}{3}
& CURL-EF & 15.3 & 15.9 & 16.0 & 16.0 & 16.0 & 16.0 \\
& CURL-LF & 18.5 & 18.8 & 18.9 & 18.8 & 18.8 & 18.8 \\
& CURL-EF\&LF & - & 18.4 & 18.4 & 18.4 & 18.4 & 18.4 \\
& CURL-EF$_n$ & 15.3 & 15.4 & 15.4 & 15.4 & 15.5 & 15.5 \\
& CURL-LF$_n$ & 18.5 & 18.5 & 18.5 & 18.6 & 18.6 & 18.6 \\
& CURL-EF\&LF$_n$ & - & 17.9 & 18.0 & 18.0 & 18.0 & 18.0 \\
\hline
\multirow{6}{*}{5}
& CURL-EF & 19.5 & 19.9 & 19.9 & 19.9 & 19.9 & 19.9 \\
& CURL-LF & 23.3 & 23.4 & 23.6 & 23.6 & 23.6 & 23.6 \\
& CURL-EF\&LF & - & 22.9 & 23.0 & 22.9 & 22.9 & 22.9 \\
& CURL-EF$_n$ & 19.5 & 19.6 & 19.7 & 19.8 & 19.8 & 19.8 \\
& CURL-LF$_n$ & 23.3 & 23.4 & 23.5 & 23.6 & 23.6 & 23.5 \\
& CURL-EF\&LF$_n$ & - & 22.8 & 22.8 & 22.9 & 22.9 & 22.9 \\
\hline
\multirow{6}{*}{10}
& CURL-EF & 25.2 & 25.4 & 25.5 & 25.5 & 25.5 & 25.5 \\
& CURL-LF & 29.9 & 29.9 & 29.9 & 29.9 & 29.9 & 29.9 \\
& CURL-EF\&LF & - & 29.3 & 29.3 & 29.3 & 29.3 & 29.3 \\
& CURL-EF$_n$ & 25.2 & 25.6 & 25.9 & 26.0 & 26.1 & 26.2 \\
& CURL-LF$_n$ & 29.9 & 30.4 & 30.5 & 30.6 & 30.6 & 30.6 \\
& CURL-EF\&LF$_n$ & - & 29.7 & 29.9 & 30.0 & 30.0 & 30.0 \\
\hline
\multirow{6}{*}{15}
& CURL-EF & 29.0 & 29.3 & 29.3 & 29.3 & 29.3 & 29.3 \\
& CURL-LF & 33.8 & 34.0 & 34.1 & 34.1 & 34.1 & 34.1 \\
& CURL-EF\&LF & - & 33.6 & 33.7 & 33.6 & 33.6 & 33.7 \\
& CURL-EF$_n$ & 29.0 & 29.7 & 29.9 & 30.1 & 30.2 & 30.3 \\
& CURL-LF$_n$ & 33.8 & 34.3 & 34.6 & 34.7 & 34.8 & 34.8 \\
& CURL-EF\&LF$_n$ & - & 34.0 & 34.3 & 34.4 & 34.5 & 34.5 \\
\hline
\multirow{6}{*}{20}
& CURL-EF & 31.0 & 31.2 & 31.2 & 31.2 & 31.2 & 31.2 \\
& CURL-LF & 36.0 & 36.2 & 36.3 & 36.3 & 36.3 & 36.3 \\
& CURL-EF\&LF & - & 35.9 & 35.9 & 35.9 & 35.9 & 35.9 \\
& CURL-EF$_n$ & 31.0 & 31.9 & 32.2 & 32.4 & 32.5 & 32.5 \\
& CURL-LF$_n$ & 36.0 & 36.6 & 36.9 & 37.0 & 37.1 & 37.1 \\
& CURL-EF\&LF$_n$ & - & 36.5 & 36.8 & 36.9 & 37.0 & 37.0 \\
\hline
\end{tabular}
\\
\end{tabular}
}
\end{center}
\end{table*}

The behavior of the methods is quite stable with respect to the three
learning scenarios, with slightly lower MAP obtained in the case of
self-taught learning.  It is evident that our \coso{} outperformed the other
methods in the state of the art included in the comparison across all
the data sets and all the scenarios considered.  Among the variants
considered, \curllf{ }demonstrated to be the best in the case of a small
number of labeled images, while \curllfn{ }obtained the best results
when more labeled data is available. Classifiers obtained on
early-fused representations performed generally worse than the \ADD{corresponding ones}
 obtained on late-fused representations, 
 but they are still uniformly better than the
original EP+LR Ensemble Projection which can be considered as their
non-cotrained version. SVMs and LapSVMs performed poorly on the
Scene-15 data set, but they outperformed EP\ADD{+LR} and some of the CURL
\ADD{variants} on the Caltech-101 data set.

\begin{figure*}[!htbp]%
\setlength{\tabcolsep}{3pt}
\begin{tabular}{cccccc}
{\includegraphics[width=0.33\columnwidth]{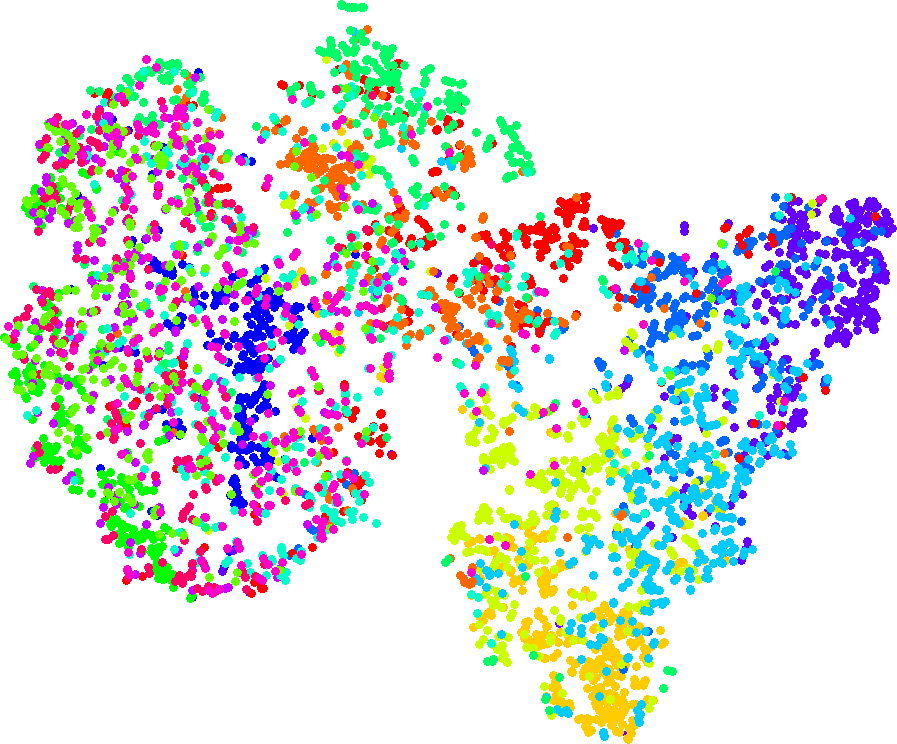}} &
{\includegraphics[width=0.32\columnwidth]{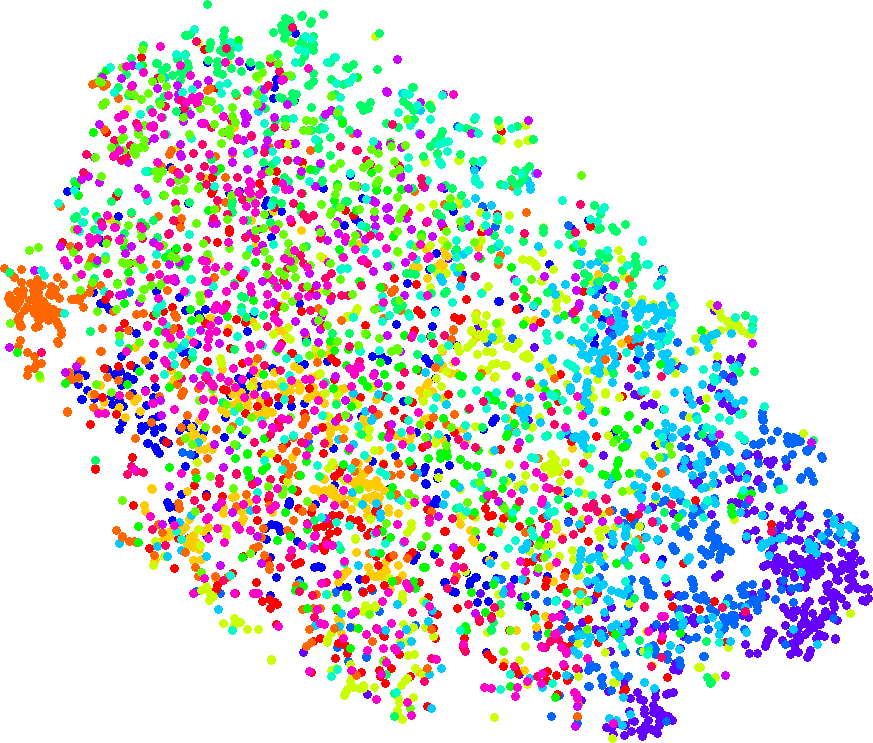}} &
{\includegraphics[width=0.31\columnwidth, angle=90]{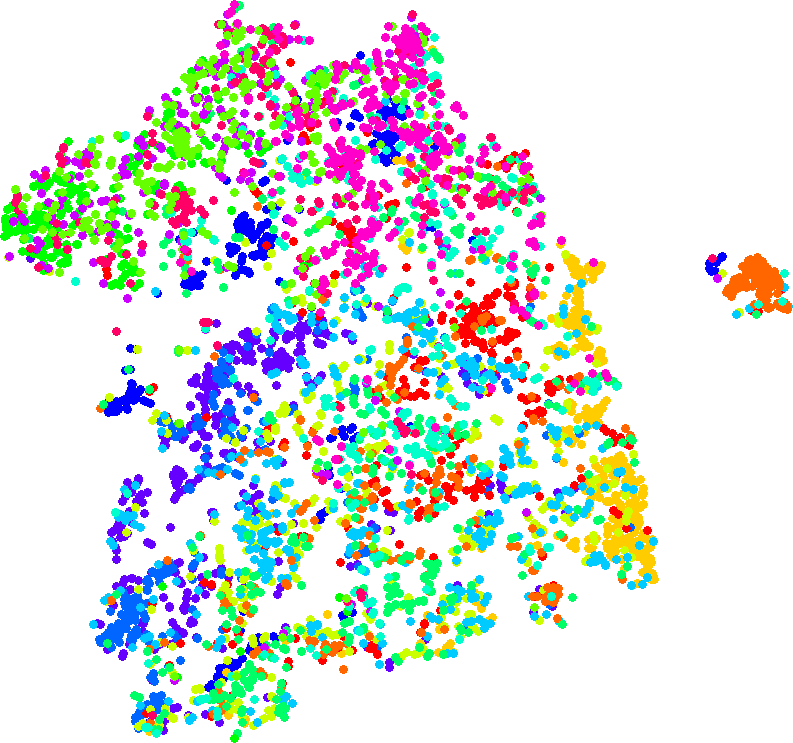}} &
{\includegraphics[width=0.31\columnwidth]{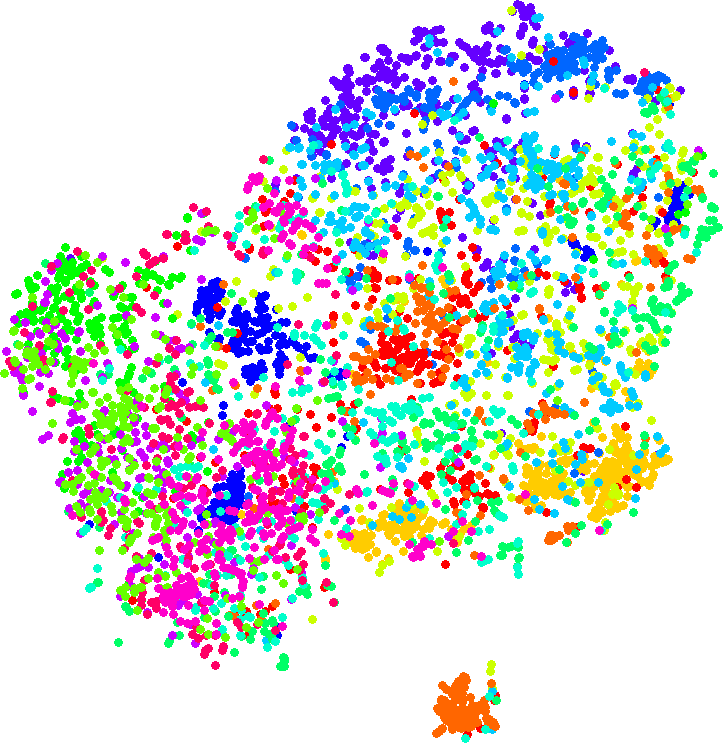}} &
{\includegraphics[width=0.32\columnwidth]{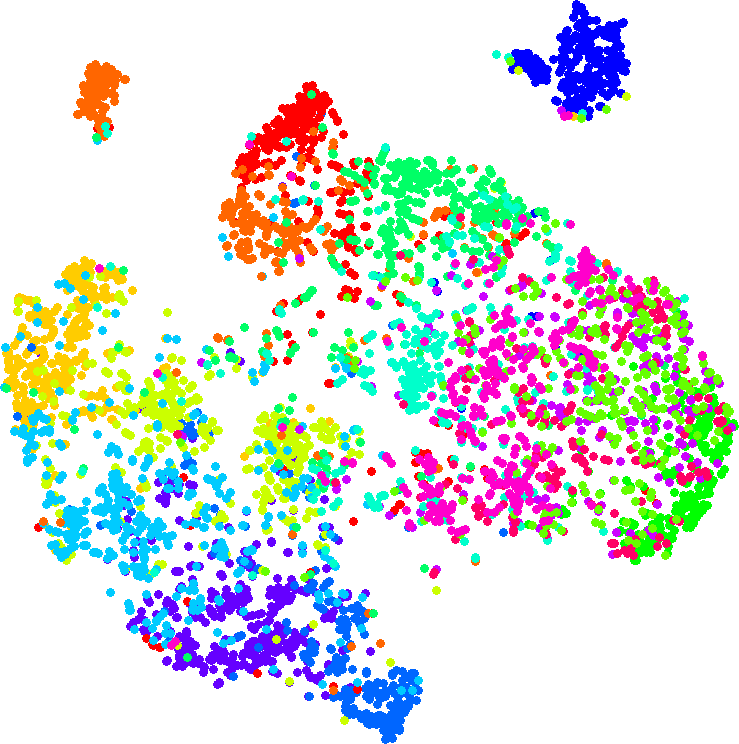}} &
{\includegraphics[width=0.32\columnwidth]{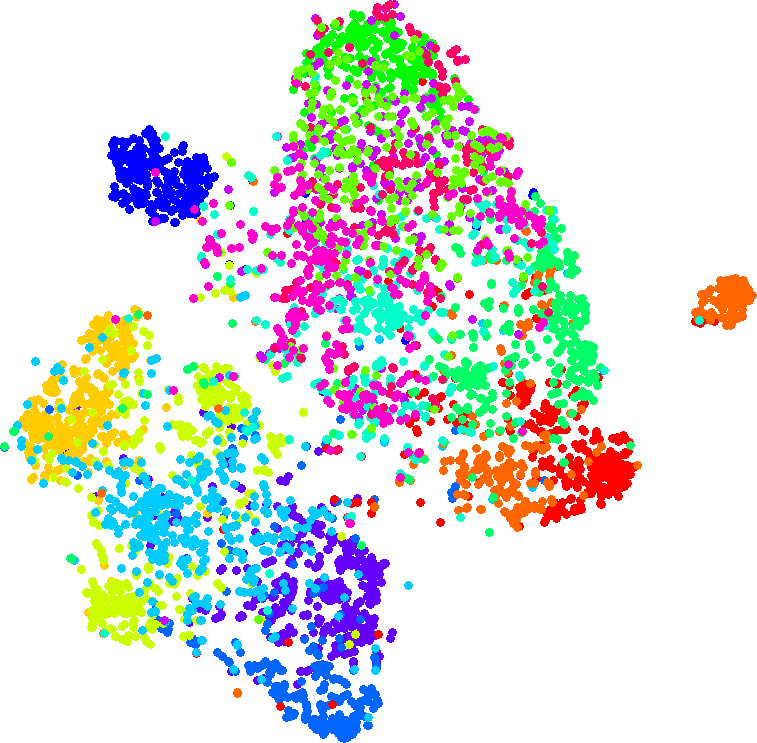}} \\
(a) GIST & (b) PHOG & (c) LBP & (d) concatenation & (e) early fusion & (f) late fusion \\
{\includegraphics[width=0.31\columnwidth]{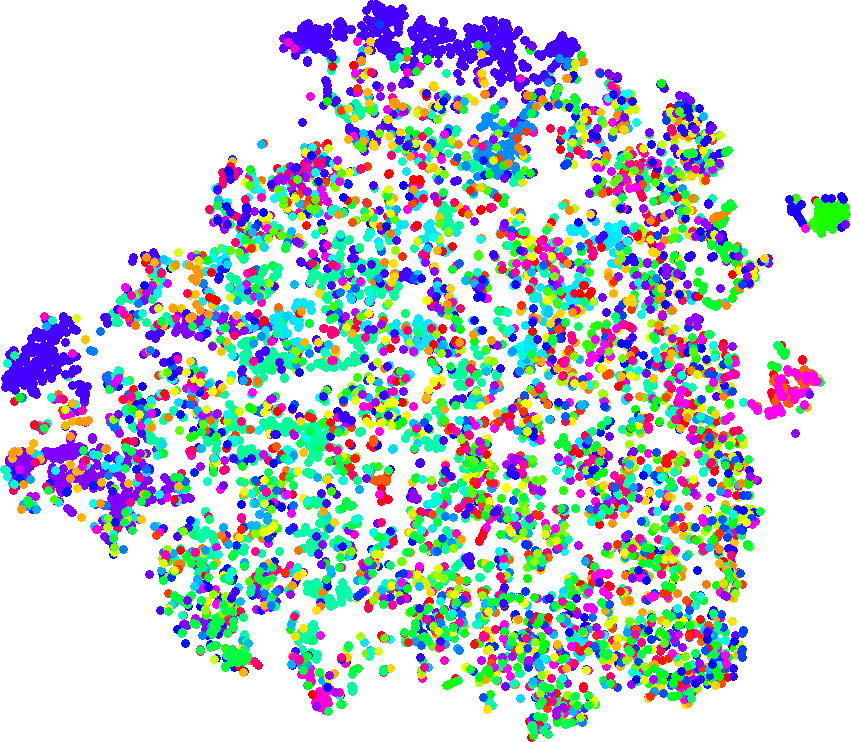}} &
{\includegraphics[width=0.29\columnwidth]{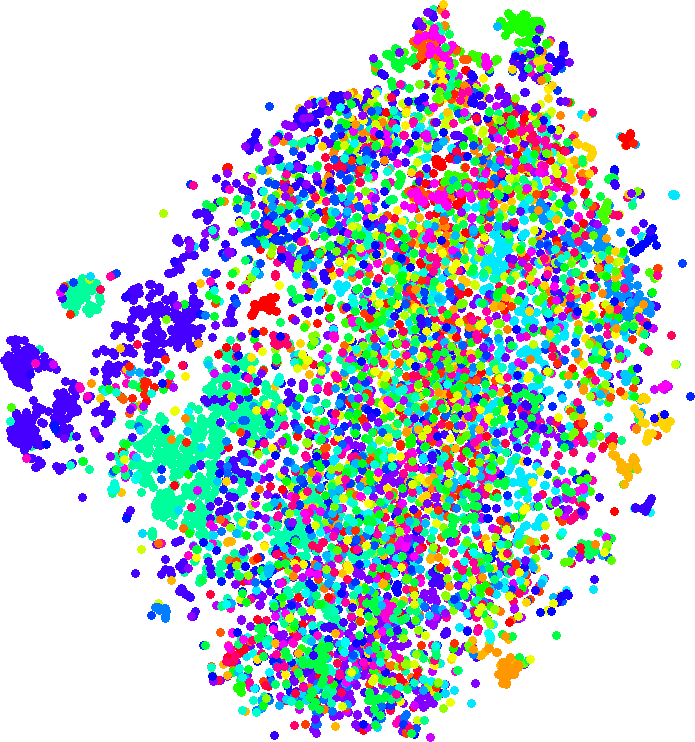}} &
{\includegraphics[width=0.29\columnwidth]{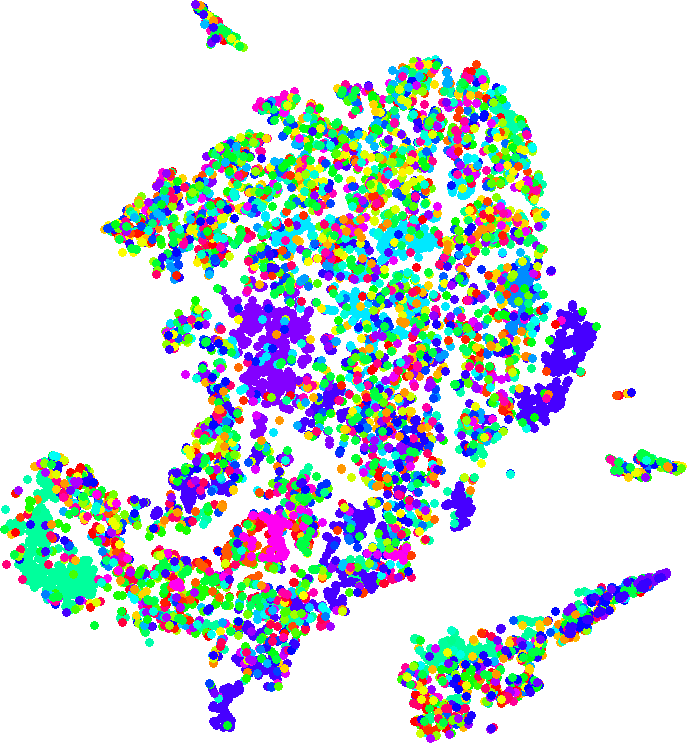}} &
{\includegraphics[width=0.29\columnwidth]{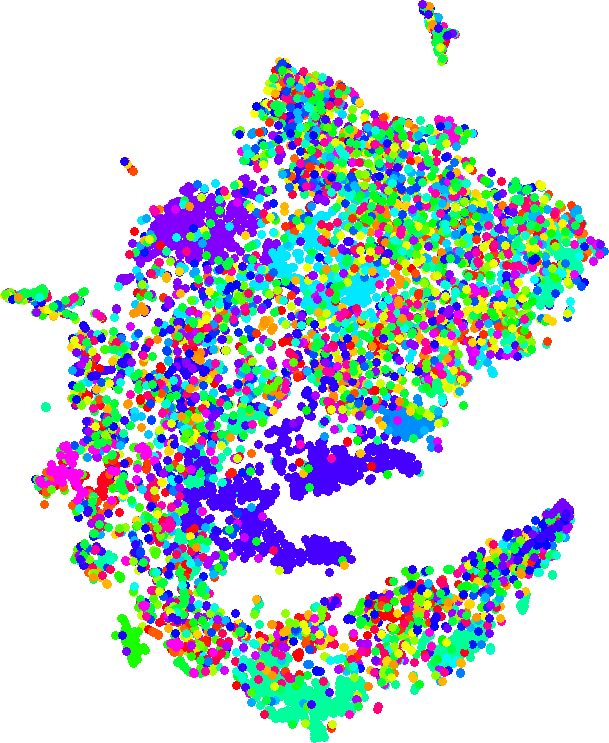}} &
{\includegraphics[width=0.29\columnwidth]{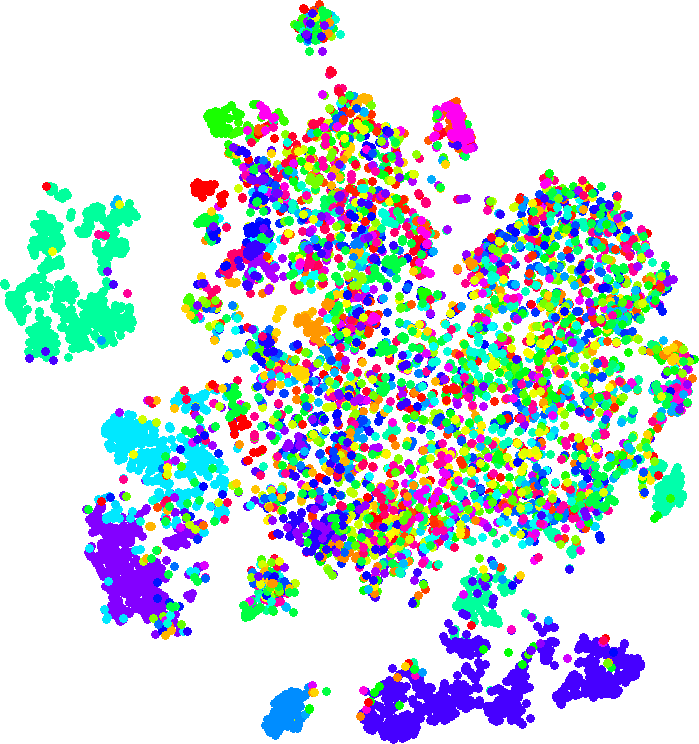}} &
{\includegraphics[width=0.30\columnwidth]{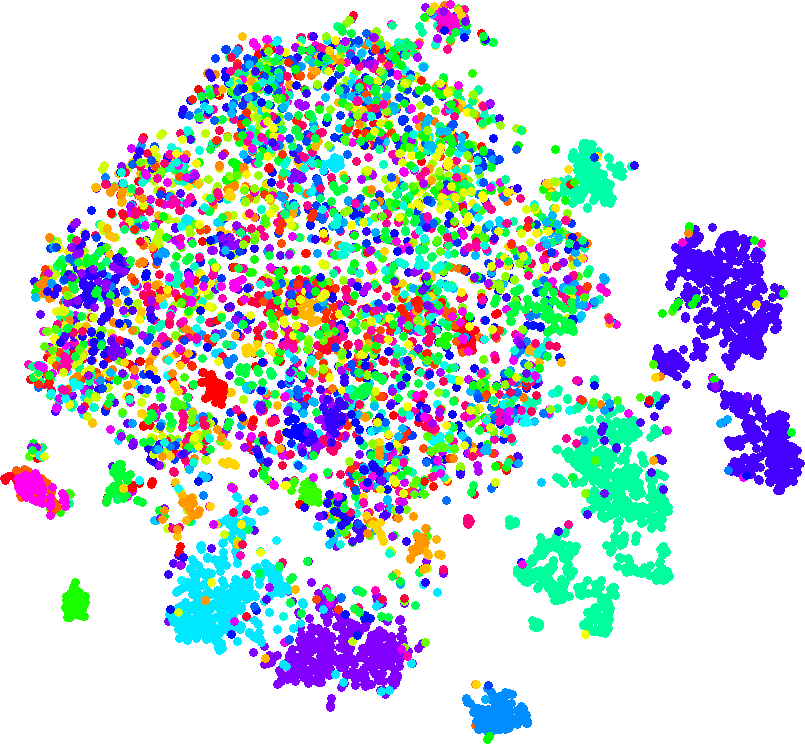}} \\
\end{tabular}
\caption{t-SNE 2D projections for the different features used.
They are relative to the Scene-15 (top row) and Caltech-101 (bottom row) data sets. Different classes are represented
  in different colors, and the same class with the same color across
  the row.}%
\label{fig:db-2DmappedFeat}%
\end{figure*}

Co-training allows to make good use of the early fusion
representations that otherwise lead to worse results than late fusion
representations.  In our opinion this happens because the two views
capture different relationships among data.  This fact is visible in
Fig.~\ref{fig:db-2DmappedFeat}, which shows 2D projections obtained by
applying the t-SNE~\cite{van2008visualizing} method to GIST, PHOG, LBP
features, their concatenation, and their learnt early- and late-fused
representations.
Unsupervised representation learning allows t-SNE to identify groups
of images of the same class.  Moreover, representations based on early
and late fusion induce different relationships among the classes.  For
instance, in the second row of Fig.~\ref{fig:db-2DmappedFeat}f the
blue and the light green classes have been placed close to each other
on the bottom right; in Fig.~\ref{fig:db-2DmappedFeat}e, instead,
the two classes are well separated.  The difference in the two
representations explains the effectiveness of co-training and
justifies the difference in performance between \curlef{ }and \curllf.

As further investigation, we also combined the two classifiers produced by the co-training procedure obtaining two other variants of CURL that we denoted as \curleflf{ }and \curleflfn.
However, in our experiments, these variants did not caused any significant
improvement when compared to \curllf.


\subsection{Performance across co-training rounds}
\label{subsec:rounds}
\begin{figure*}[!htbp]%
\centering
\renewcommand{\tabcolsep}{0cm}
\scriptsize
\begin{tabular}{ccc}
Scene-15 inductive & Scene-15 transductive & Scene-15 self-taught\\ 
\includegraphics[width=0.68\columnwidth]{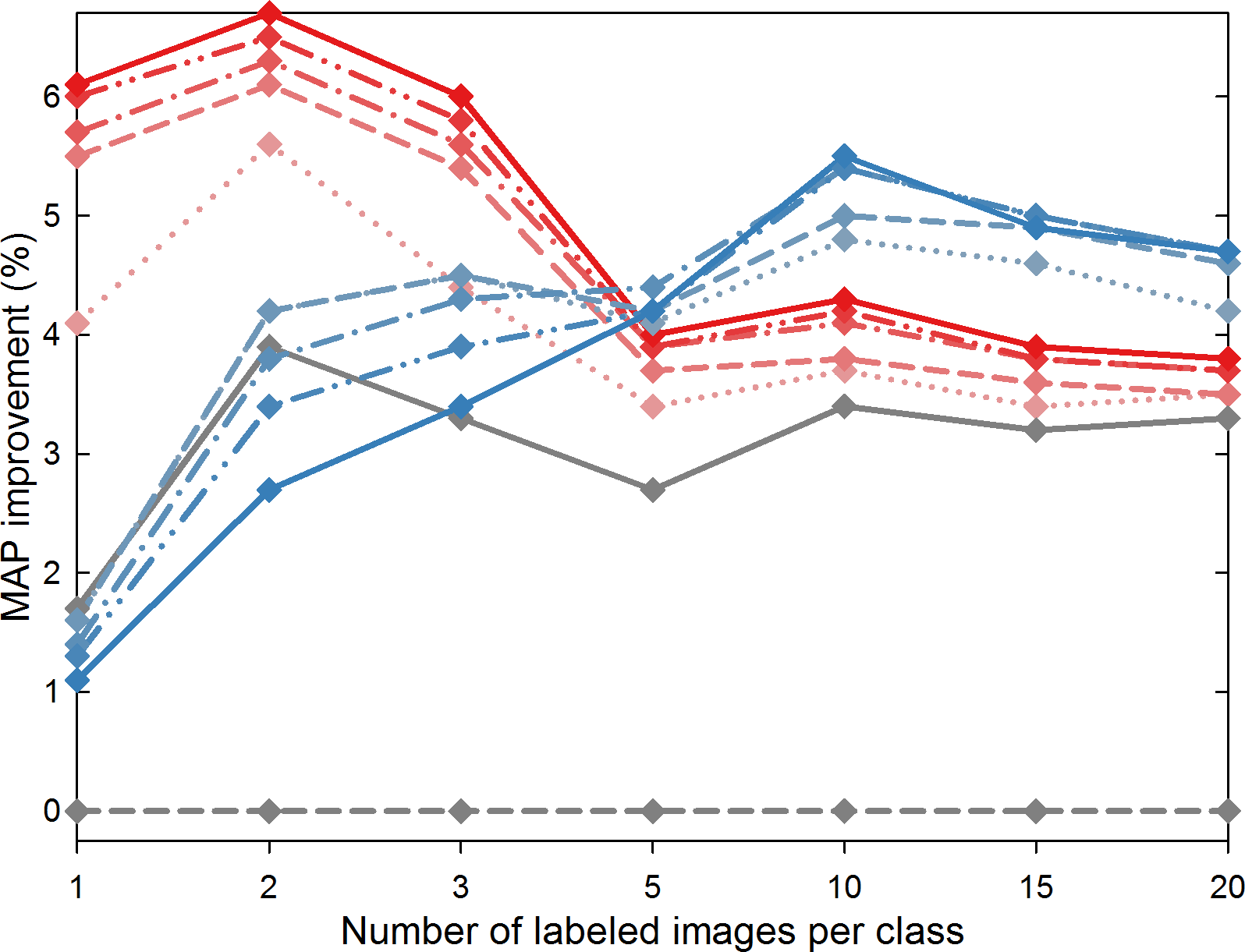} &
\includegraphics[width=0.68\columnwidth]{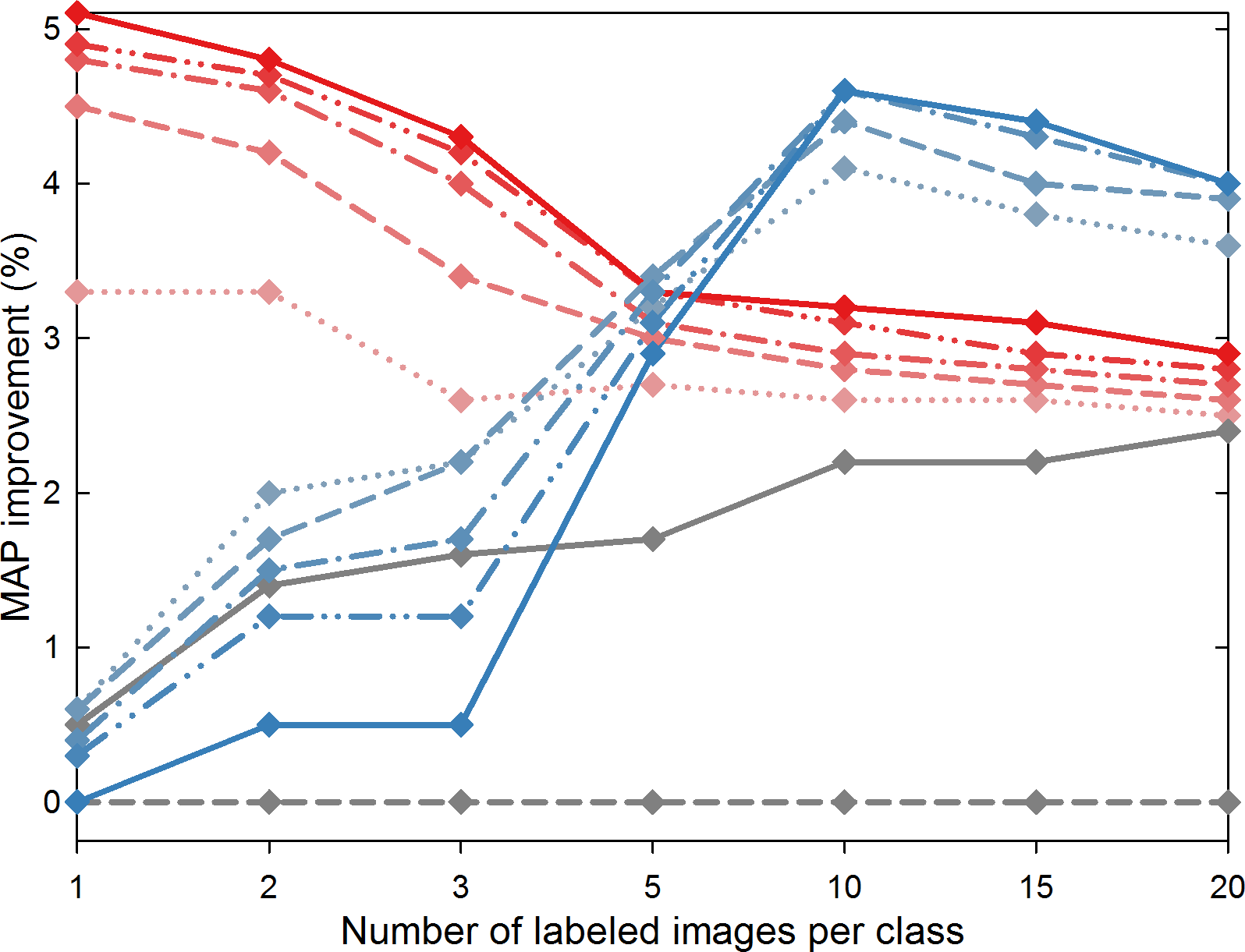} &
\includegraphics[width=0.68\columnwidth]{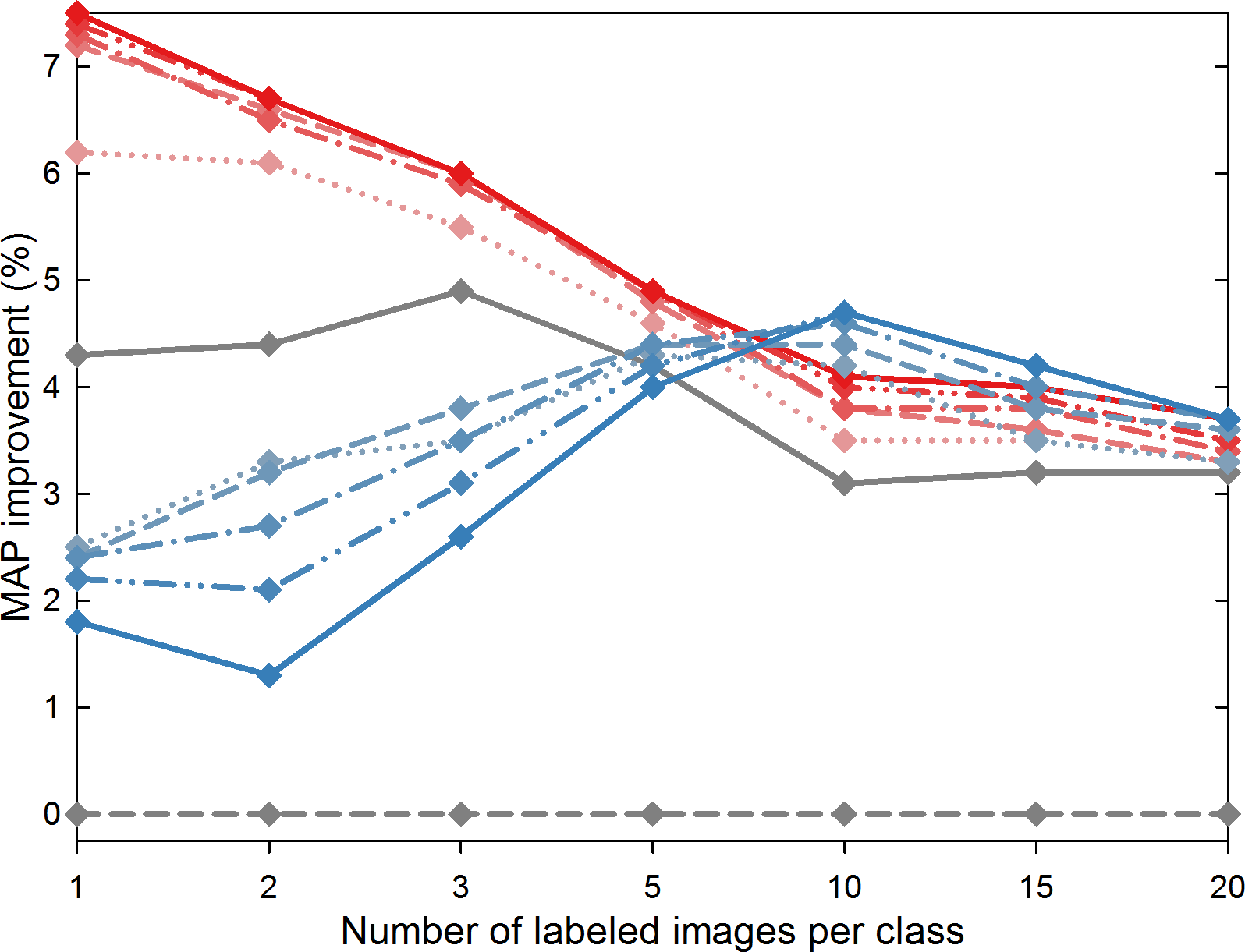} \\
\\
Caltech-101 inductive & Caltech-101 transductive & Caltech-101 self-taught\\ 
\includegraphics[width=0.68\columnwidth]{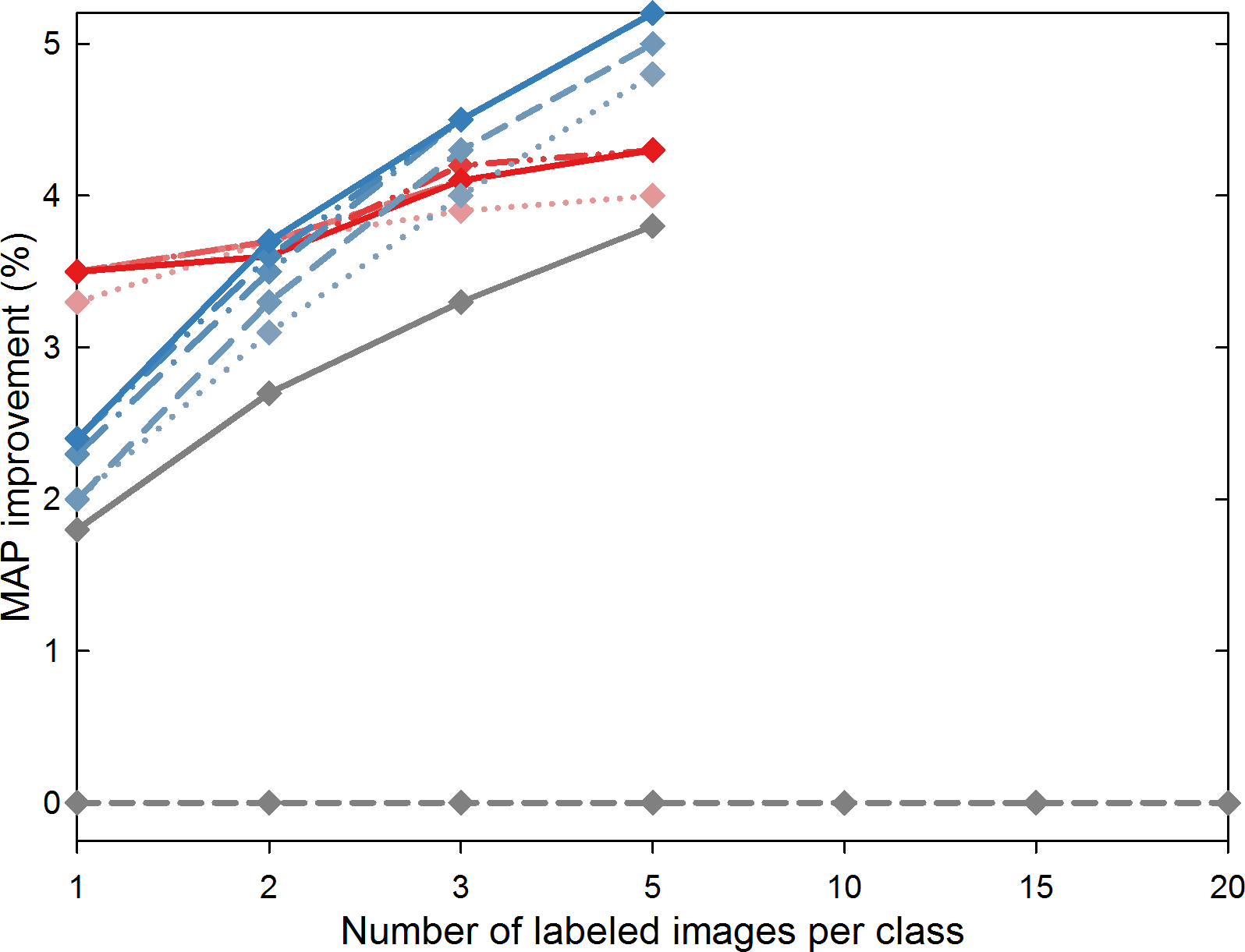} &
\includegraphics[width=0.68\columnwidth]{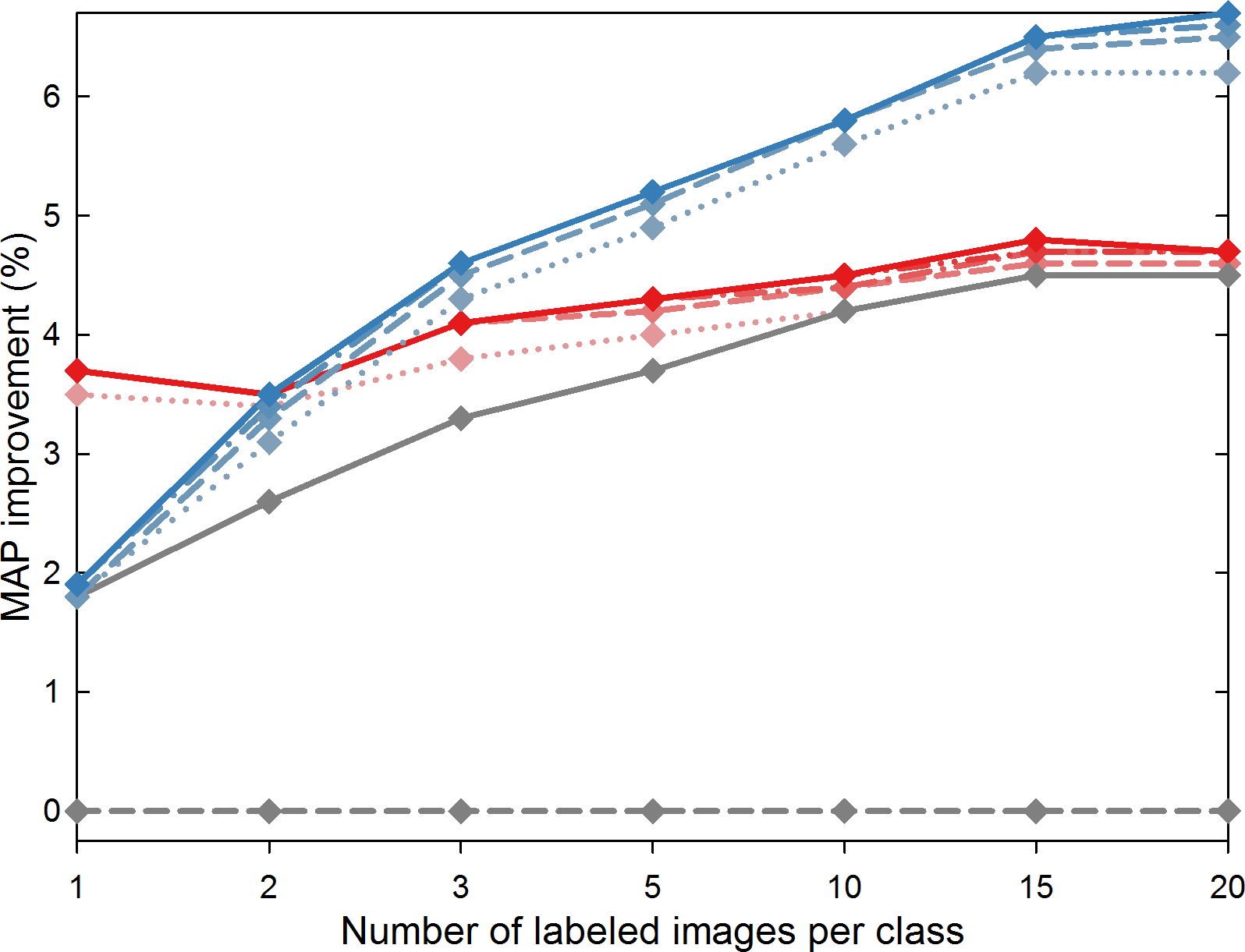} &
\includegraphics[width=0.68\columnwidth]{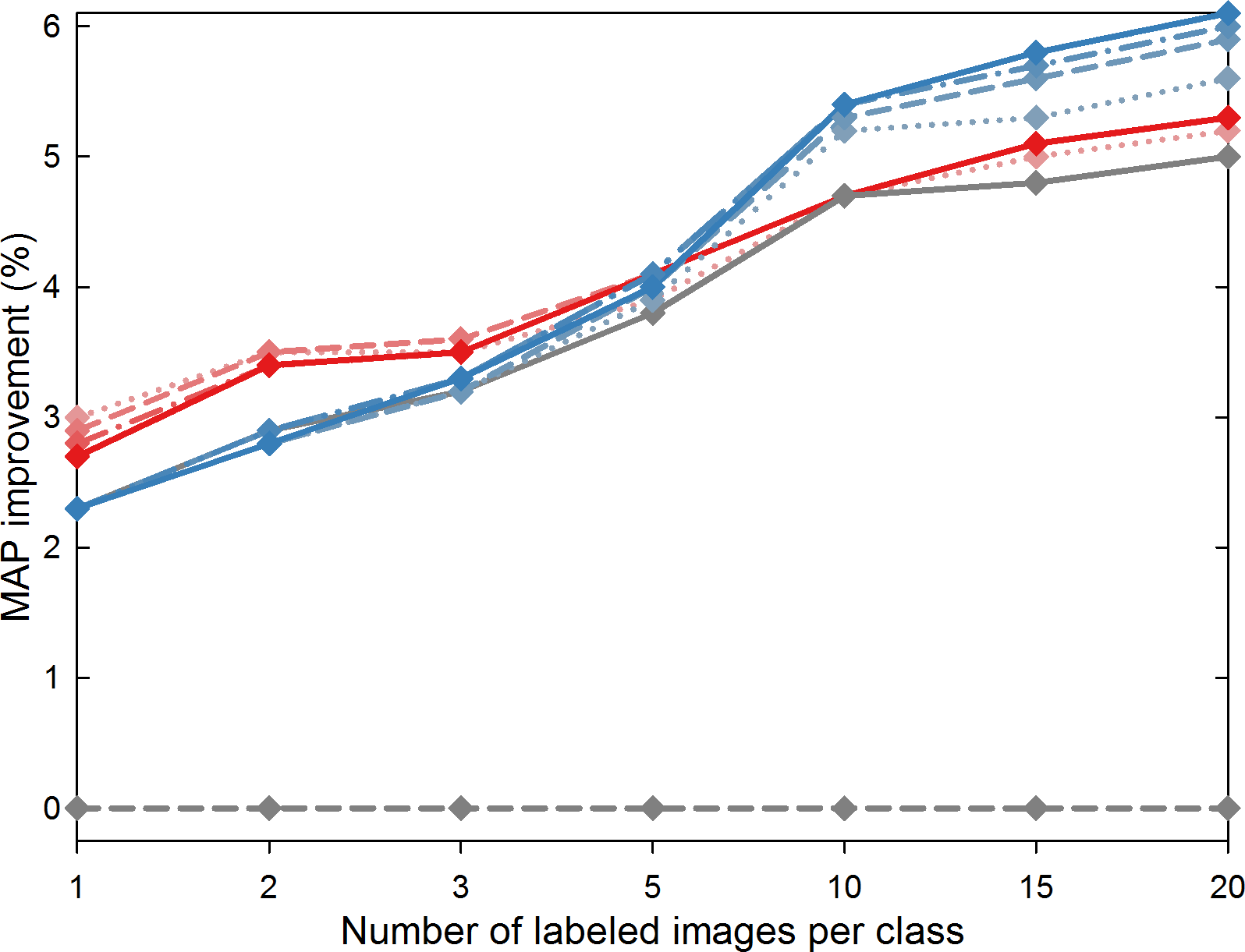} \\
\\
\multicolumn{3}{c}{\includegraphics[width=1.30\columnwidth]{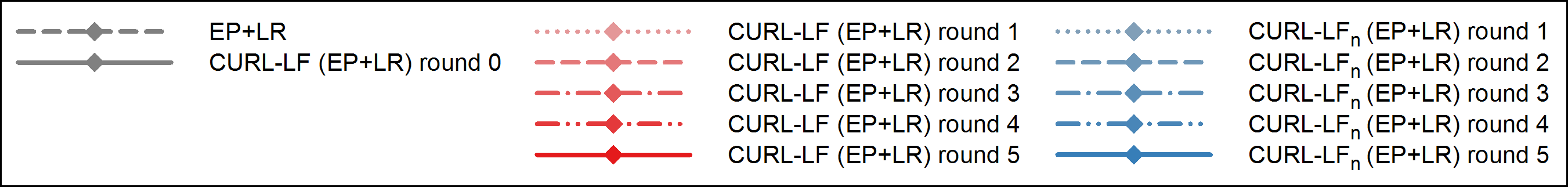}} \\
\\
\end{tabular}
\caption{Performance obtained by \curllf{ }and \curllfn{ }varying the
  number of co-training rounds.  
  Performance are reported in terms of MAP improvement
  with respect to Ensemble Projection.  Due to the small cardinality
  of some classes, inductive learning on the Caltech-101 has been
  limited to five labeled images per class.}
\label{fig:improvVAR}
\end{figure*}

Here we analyze \ADD{in more details} the performance of our \coso{ }across the five 
co-training rounds. Results are reported in Fig.~\ref{fig:improvVAR} with lines of increasing color saturation corresponding to rounds one to five. \curllf{ }is reported in red lines, while \curllfn{ }in blue. 
Results are reported in terms of MAP
improvements with respect to EP\ADD{+LR}, which, we recall, corresponds to
\curlef{ }with zero co-training rounds. For \curllf, performances always
increase with the number of rounds. For \curllfn, this is not true on
the Scene-15 data set with a small number of labeled examples.  In
\curllfn{ }each round of co-training adds all the promising unlabeled
samples, with a high chance of including some of them with the wrong
pseudo-label. This may result in a `concept drift', with the
classifiers being pulled away from the concepts represented by the
labeled examples.  This risk is lower on the Caltech-101 (which tends
to have more homogeneous classes than Scene-15) and when there are
more labeled images. The original \curllf{ }is more conservative, since
each of its co-training rounds adds a single image per class.  As a
result, increasing the rounds usually increases MAP and never decreases it by an 
appreciable amount.

We observed the same behavior for \curlef{ }and \curlefn. We omit the relative figures for sake of brevity. 

The plots confirm that \curllf{ }is better suited for small sets of
labeled images, while \curllfn{ }is to be preferred when more labeled
examples are available. The representation learned from late fused
features explains part of the effectiveness of CURL. In fact, even
\curllf{ }without co-training (zero rounds) outperforms the baseline
represented by Ensemble Projection.



\subsection{Leveraging CNN features in CURL}
\label{subsec:cnn}
In this further experiment we want to test if the proposed classification \coso{ }works when more powerful features are used. Recent results indicate that the generic descriptors extracted from pre-trained Convolutional Neural Networks (CNN) are able to obtain consistently superior results compared to the highly tuned state of the art systems in all the visual classification tasks on various datasets \cite{razavian2014cnn}. 
We extract a 4096-dimensional feature vector from each image using the Caffe \cite{jia2014caffe} implementation of the deep CNN described by Krizhevsky et al. \cite{krizhevsky2012imagenet}.  
The CNN was discriminatively trained on a large dataset (ILSVRC 2012) with image-level annotations to classify images into 1000 different classes. 
Briefly, a mean-subtracted $227 \times 227$ RGB image is forward propagated through five convolutional layers and two fully connected layers. Features are obtained by extracting activation values of the last hidden layer. More details about the network architecture can be found in \cite{krizhevsky2012imagenet,jia2014caffe}.

\begin{figure}[tbp]%
\centering
\begin{tabular}{cc}
\footnotesize{Scene-15} & \footnotesize{Caltech-101}\\
\includegraphics[width=0.4\columnwidth]{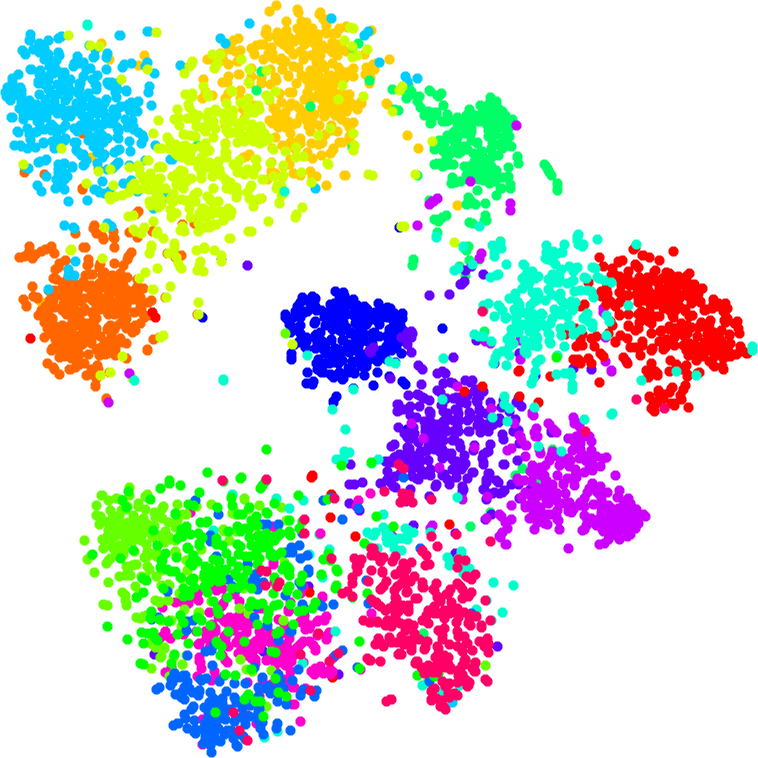} &
\includegraphics[width=0.45\columnwidth]{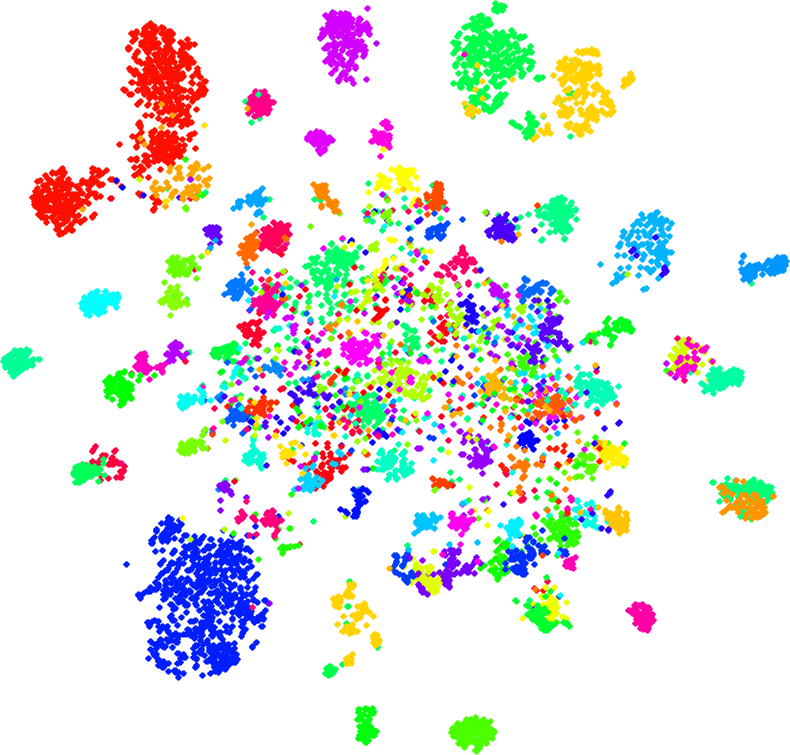} \\
\end{tabular}
\caption{2D projections for the CNN features on the two data sets used: Scene-15 (left) and Caltech-101 (right). Different classes are represented in different colors.}%
\label{fig:tsneCNN}%
\end{figure}

We leverage the CNN features in CURL using them as a fourth feature in addition to the three used in Section \ref{sec:experiments}. The discriminative power of these CNN features alone can be seen in Fig. \ref{fig:tsneCNN}, where their 2D projections obtained applying the t-SNE~\cite{van2008visualizing} method are reported. 

\begin{figure*}[ht]%
\centering
\begin{tabular}{cc}
\footnotesize{Scene-15 transductive} & \footnotesize{Caltech-101 transductive}\\
\includegraphics[width=0.80\columnwidth]{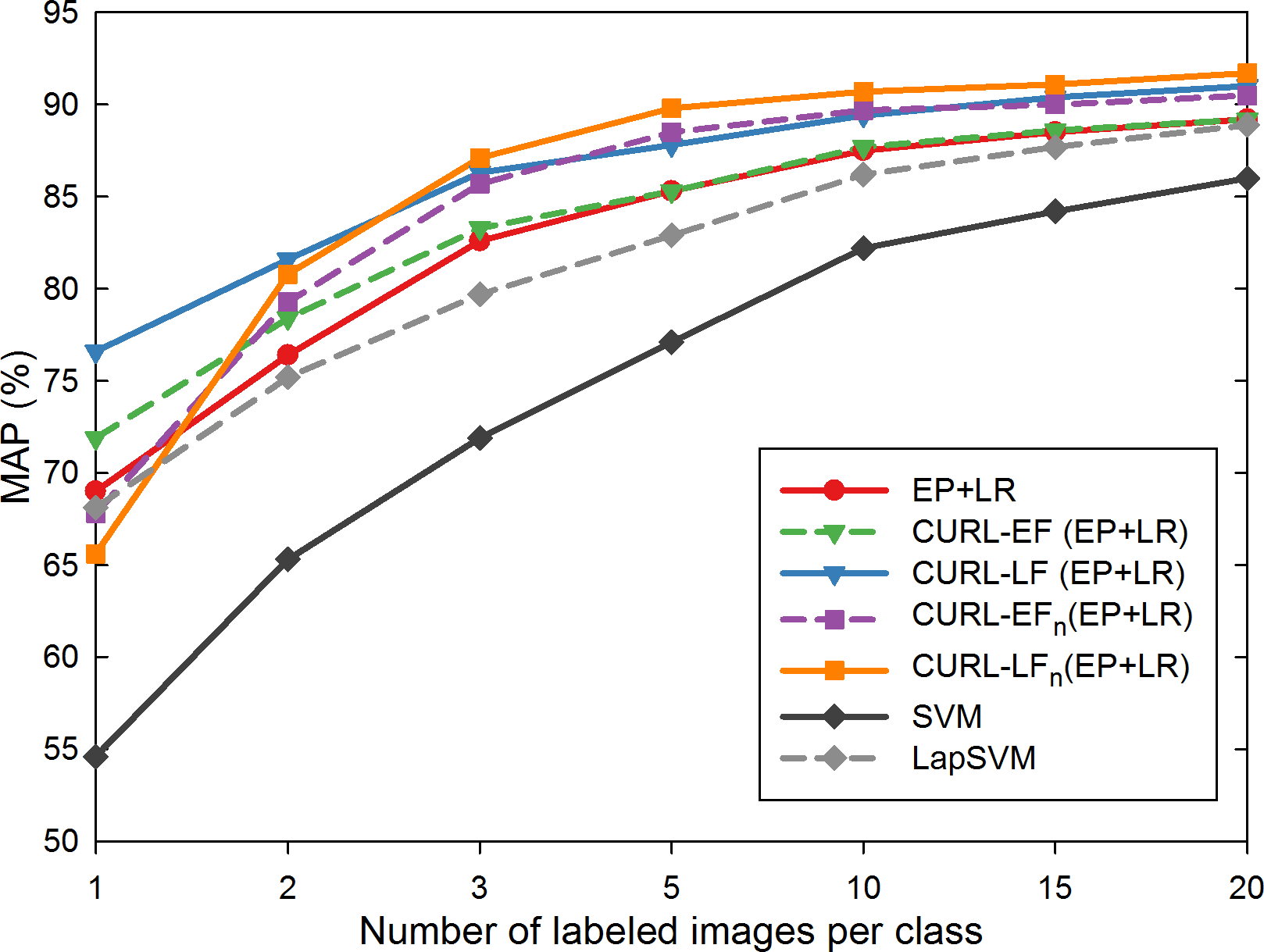} &
\includegraphics[width=0.80\columnwidth]{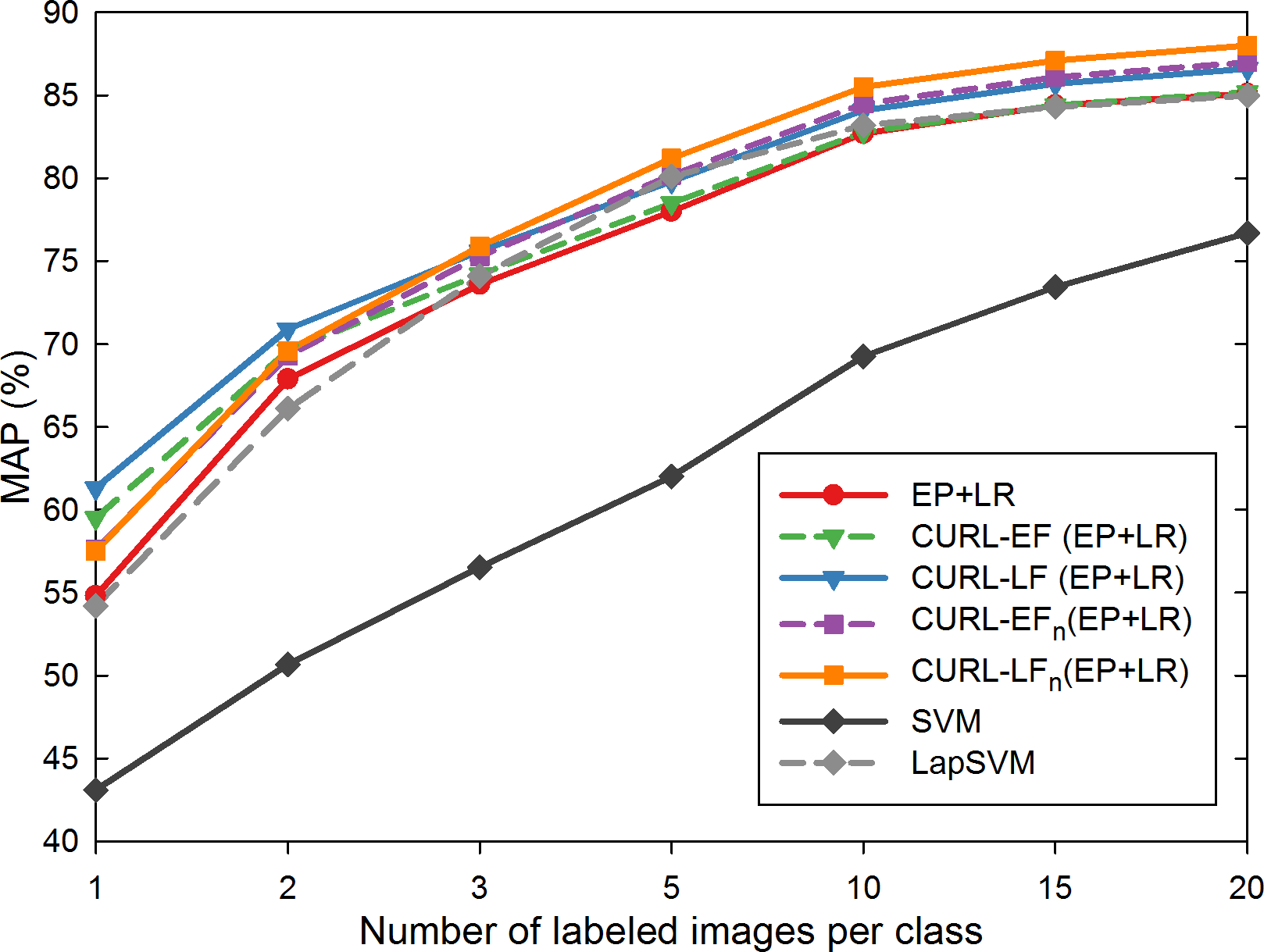} \\
\end{tabular}
\caption{Mean Average Precision (MAP) varying the number of labeled images per class, obtained on the Scene-15 data set (left), and on the Caltech-101
data set (right). Results are obtained using GIST, PHOG, LBP and CNN features.}
\label{fig:epcnn}
\end{figure*}

The experimental results using the four features, are reported in Fig. \ref{fig:epcnn}, for both the Scene-15 and Caltech-101 data sets. We report the results in the transductive scenario only. It can be seen that the results using the four features are significantly better than those using only three features mainly due to the discriminative power of the CNN features. Furthermore, the CURL variants achieve better results than the baselines. This suggests that CURL is able to effectively leverage both low/mid level features as LBP, PHOG and GIST, and  more powerful features as CNN.

\subsection{Second \emb{ }of CURL using LapSVM}
\label{subsec:curllap}
\begin{figure*}[!htbp]%
\centering
\begin{tabular}{cc}
\footnotesize{Scene-15 transductive} & \footnotesize{Caltech-101 transductive}\\
\includegraphics[width=0.80\columnwidth]{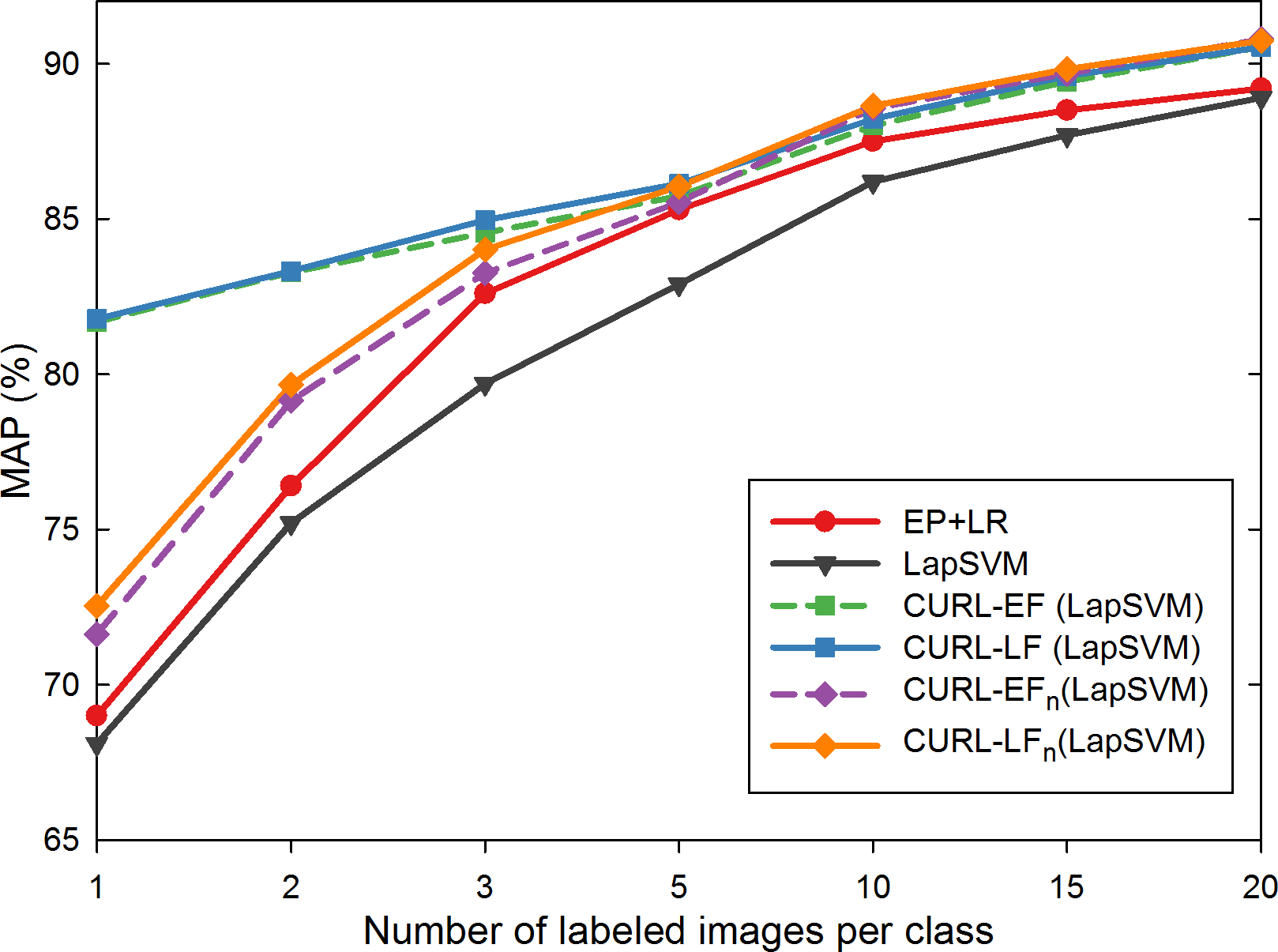} &
\includegraphics[width=0.80\columnwidth]{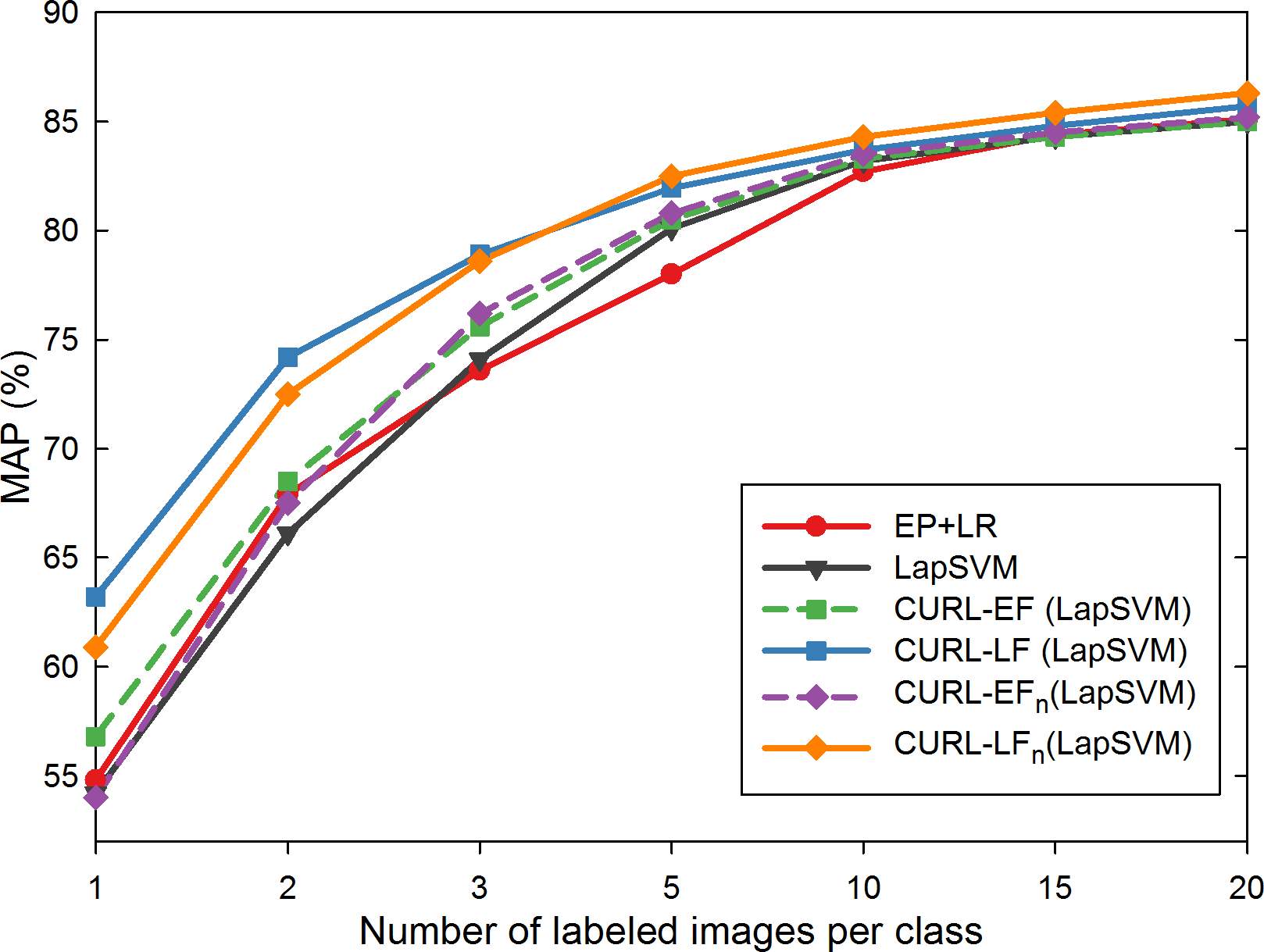} \\
\end{tabular}
\caption{Mean Average Precision (MAP) varying the number of labeled images per class, obtained on the Scene-15 data set (left), and on the Caltech-101 data set (right). Results are obtained using GIST, PHOG, LBP and CNN features.}%
\label{fig:curllap}%
\end{figure*}

\ADD{
In this Section we want to evaluate the CURL performance in a different \emb. Specifically, we substitute the EP and LR components with LapSVM-based ones. 
In the LapSVM, first, an unsupervised geometrical deformation of the feature kernel is performed. This deformed kernel is then used for classification by a standard SVM thus by-passing an explicit definition of a new feature representation. In this CURL \emb{ }we exploit the unsupervised step as surrogate of the URL component, and SVM as C component. 
The EF view is obtained concatenating the GIST, PHOG, LBP and CNN features and generating the corresponding kernel, while the LF one is obtained by a linear combination of the four kernels computed on each feature. This is similar to what is done in multiple kernel learning \cite{gonen2011multiple}. Due to its performance in the previous experiments, the $\chi^2$ kernel is used for both views. 
The experimental results on the Scene-15 and Caltech-101 data sets in the transductive scenario, are reported in Fig. \ref{fig:curllap}. We named the variants of this CURL \emb{ }by adding the suffix (LapSVM). It can be seen that the behavior of the different methods is the same of the previous plots, with the LapSVM-based CURL outperforming the standard LapSVM. The plots confirm that \curllaplf{ }is better suited for small sets of labeled images, while \curllaplfn{ }is to be preferred when more labeled examples are available.}

\begin{figure*}[ht]%
\begin{overpic}[height=4.1in]{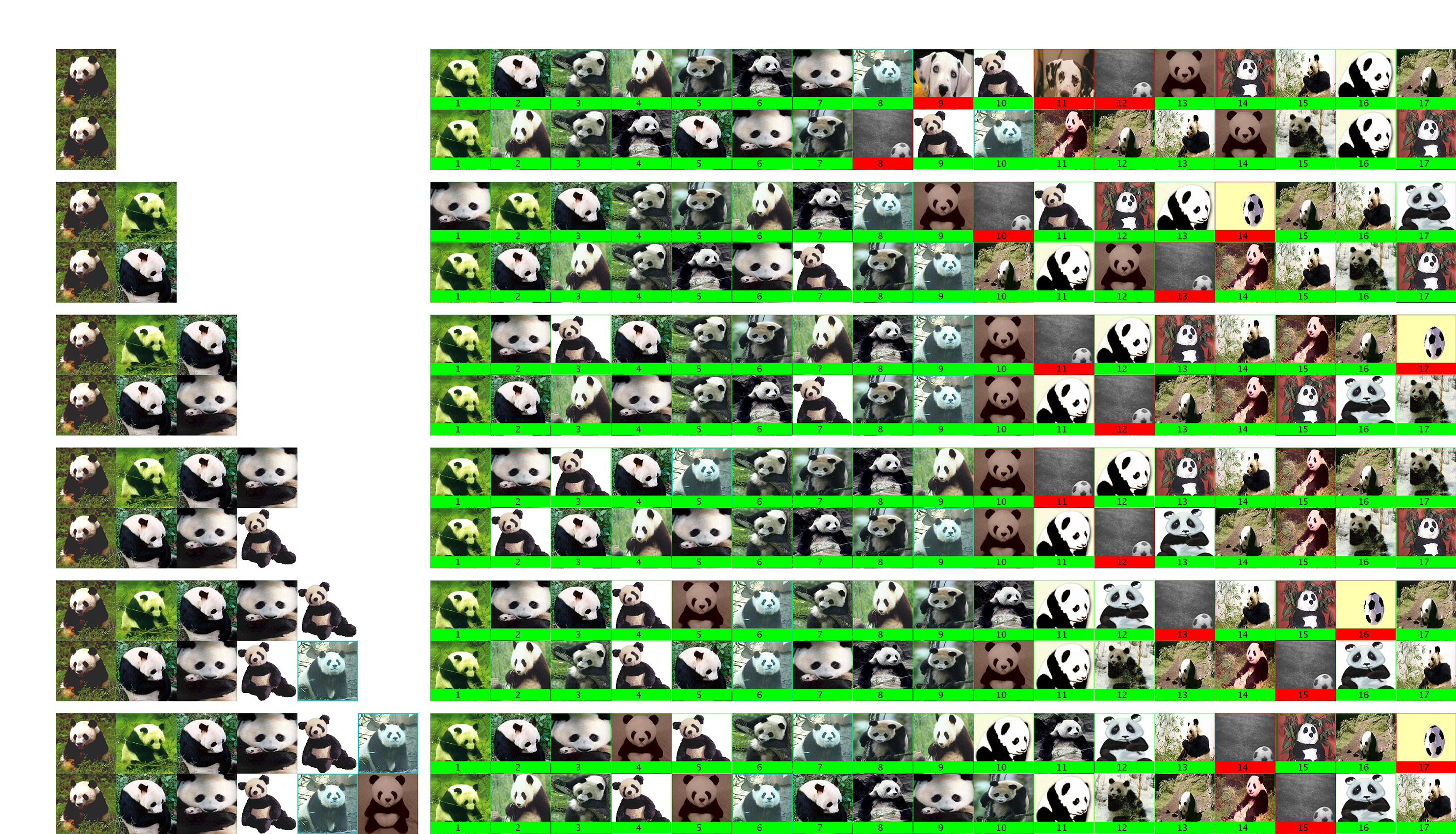}
\put(0,47){\rotatebox{90}{\scriptsize round 0}}
\put(1.5,51){\scriptsize EF}
\put(1.5,47){\scriptsize LF}
\put(0,38){\rotatebox{90}{\scriptsize round 1}}
\put(1.5,42){\scriptsize EF}
\put(1.5,38){\scriptsize LF}
\put(0,29){\rotatebox{90}{\scriptsize round 2}}
\put(1.5,33){\scriptsize EF}
\put(1.5,29){\scriptsize LF}
\put(0,20){\rotatebox{90}{\scriptsize round 3}}
\put(1.5,24){\scriptsize EF}
\put(1.5,20){\scriptsize LF}
\put(0,11){\rotatebox{90}{\scriptsize round 4}}
\put(1.5,15){\scriptsize EF}
\put(1.5,11){\scriptsize LF}
\put(0,2){\rotatebox{90}{\scriptsize round 5}}
\put(1.5,6){\scriptsize EF}
\put(1.5,2){\scriptsize LF}
\put(4,55){\scriptsize Training images}
\put(30,55){\scriptsize Test images}
\end{overpic}
\caption{Qualitative results of the proposed \coso{ }for the `Panda' class of the Caltech-101 data set over five co-training rounds. Train images are on the left, the first 17 test images, ordered by decreasing classification confidence are on the right. Test images from 18 to 40 are reported in Fig.~\ref{fig:quali40B}.}
\label{fig:quali40A}
\end{figure*}

\begin{figure*}[ht]%
\begin{overpic}[height=4.1in]{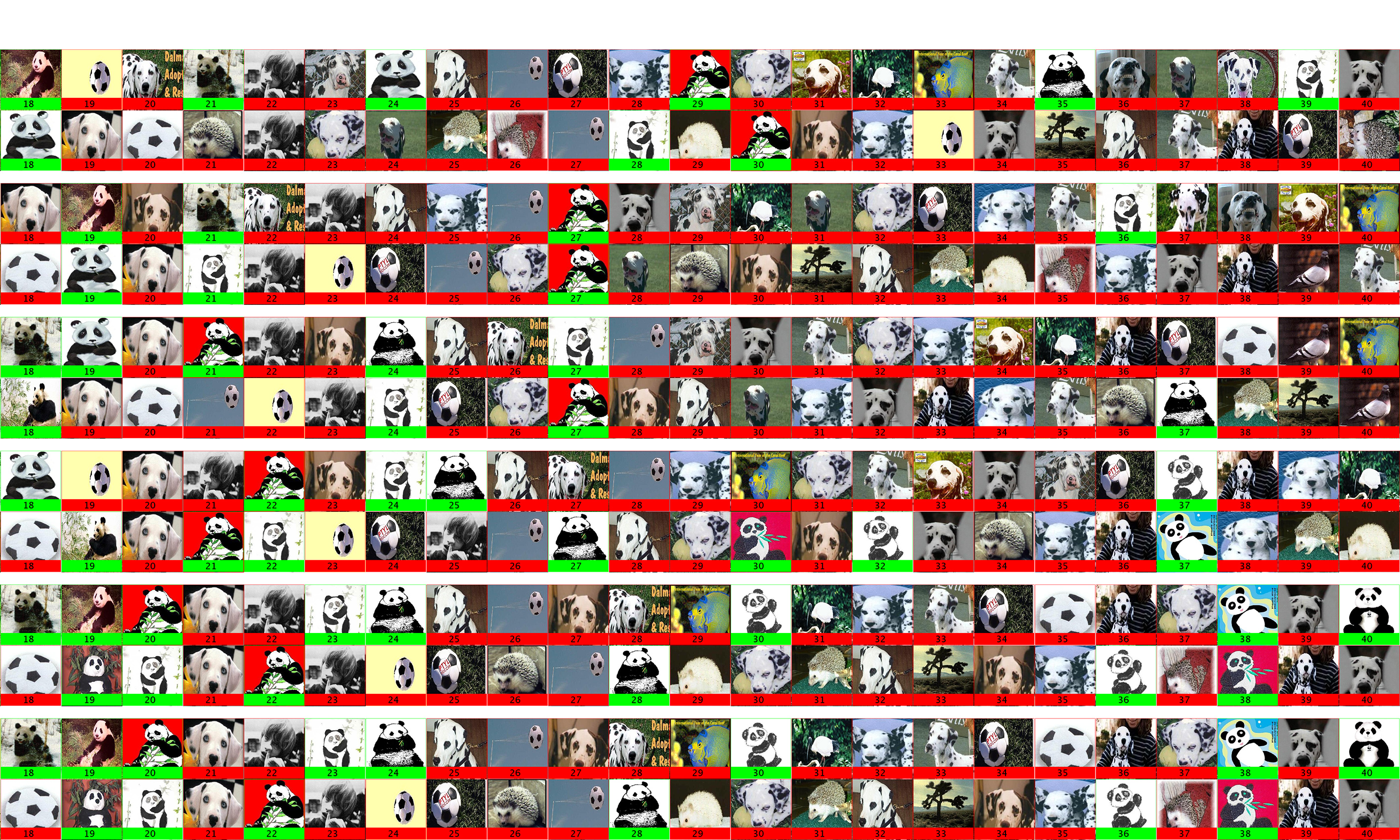}
\put(1,58){\scriptsize Test images}
\end{overpic}
\caption{Qualitative results of the proposed \coso{ }for the `Panda' class of the Caltech-101 data set over five co-training rounds. The images are ordered by decreasing classification confidence. Training image, and test images from 1 to 17 are reported in Fig.~\ref{fig:quali40A}.}
\label{fig:quali40B}
\end{figure*}

In Fig. \ref{fig:quali40A} and \ref{fig:quali40B} 
qualitative results for the `Panda' class of the Caltech-101 data set are reported: the results are relative to the case in which a single instance is available for training and one single example is added at each co-training round (i.e. each pair of rows correspond to \curllapef{ }and \curllaplf{ }respectively). The left part of Fig.~\ref{fig:quali40A} contains the training examples that are added by the \curllapef{ }and \curllaplf{ }at each co-training round, while the right part and Fig.~\ref{fig:quali40B} contain the first 40 test images ordered by decreasing classification confidence. Samples belonging to the current class are surrounded by a green bounding box, while a red one is used for samples belonging to other classes.

In the sets of training images, it is possible to see that after the first co-training round, \curllaplf{ }selects new examples to add to the training set, while \curllapef{ }adds examples seleted by \curllaplf{ }in the previous round. This is a pattern that we found to occur also in other categories when very small training sets are used. 

In the sets of test images, it is possible to see that more and more positive images are recovered. Moreover, we can see how the images belonging to the correct class tends to be classified with increasing confidence and move to the left, while the confidences of images belonging to other classes decrease and are pushed to the right.

\subsection{Large scale experiment}	
\label{subsec:ilvsrc}

In this experiment we want to test the proposed classification \coso{ }on a large scale data set, namely the ILSVRC 2012 which contains a total of 1000 different classes. The experiment is run on the ILSVRC 2012 validation set since the training set was used to learn the CNN features. The ILSVRC 2012 validation set, which contains a total of 50 images for each class, has been randomly divided into a training and a test set containing each 25 images per class. Again, different numbers of training images per class were tested (i.e. 1, 2, 3, 5, 10, and 20). The second \emb{ } of CURL is used in this experiment. 

\begin{figure*}[!htbp]%
\centering
\begin{tabular}{cc}
\footnotesize{ILSVRC 2012 transductive}  & \footnotesize{ILSVRC 2012 transductive}  \\
\includegraphics[width=0.80\columnwidth]{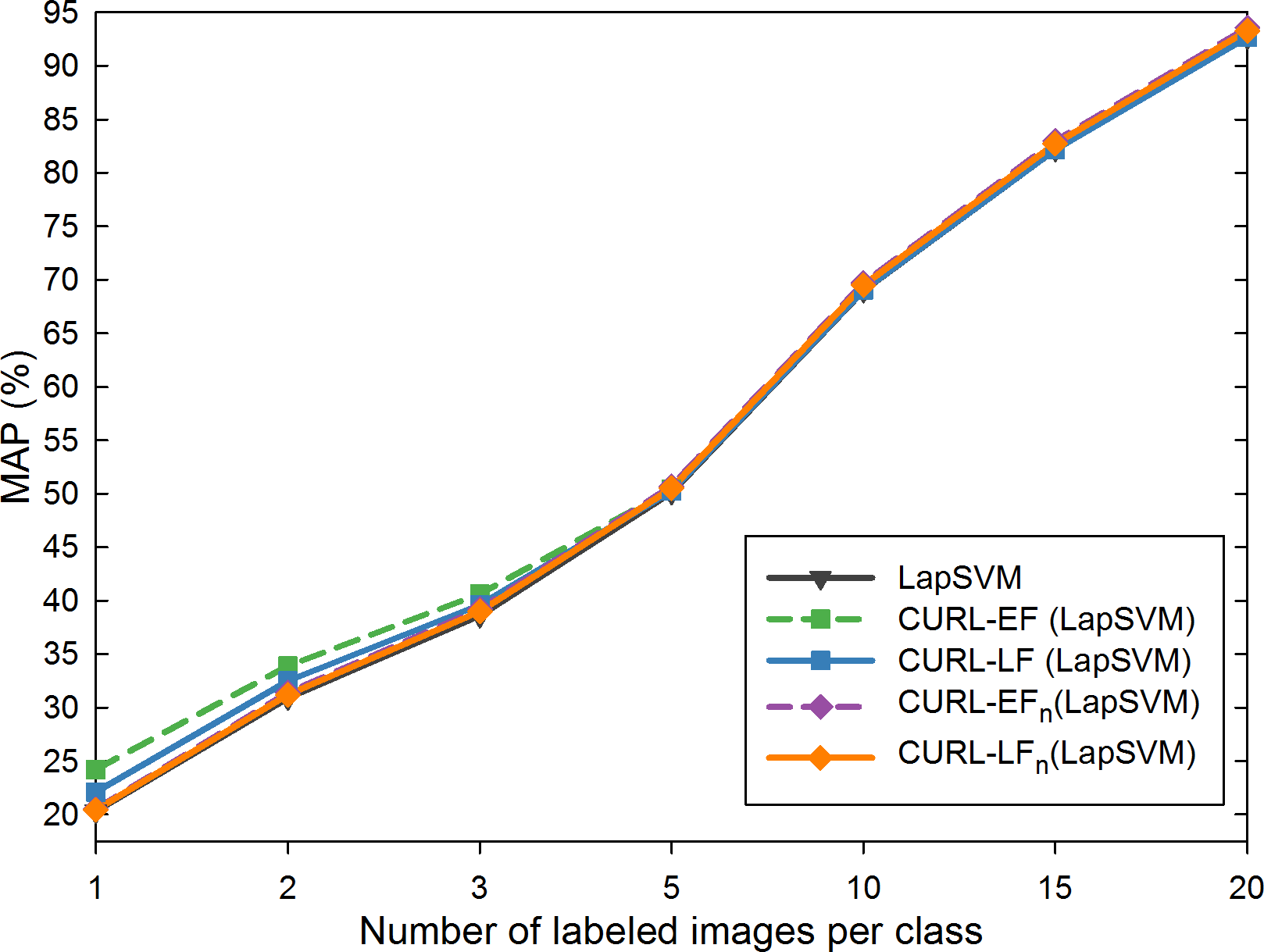} &
\includegraphics[width=0.80\columnwidth]{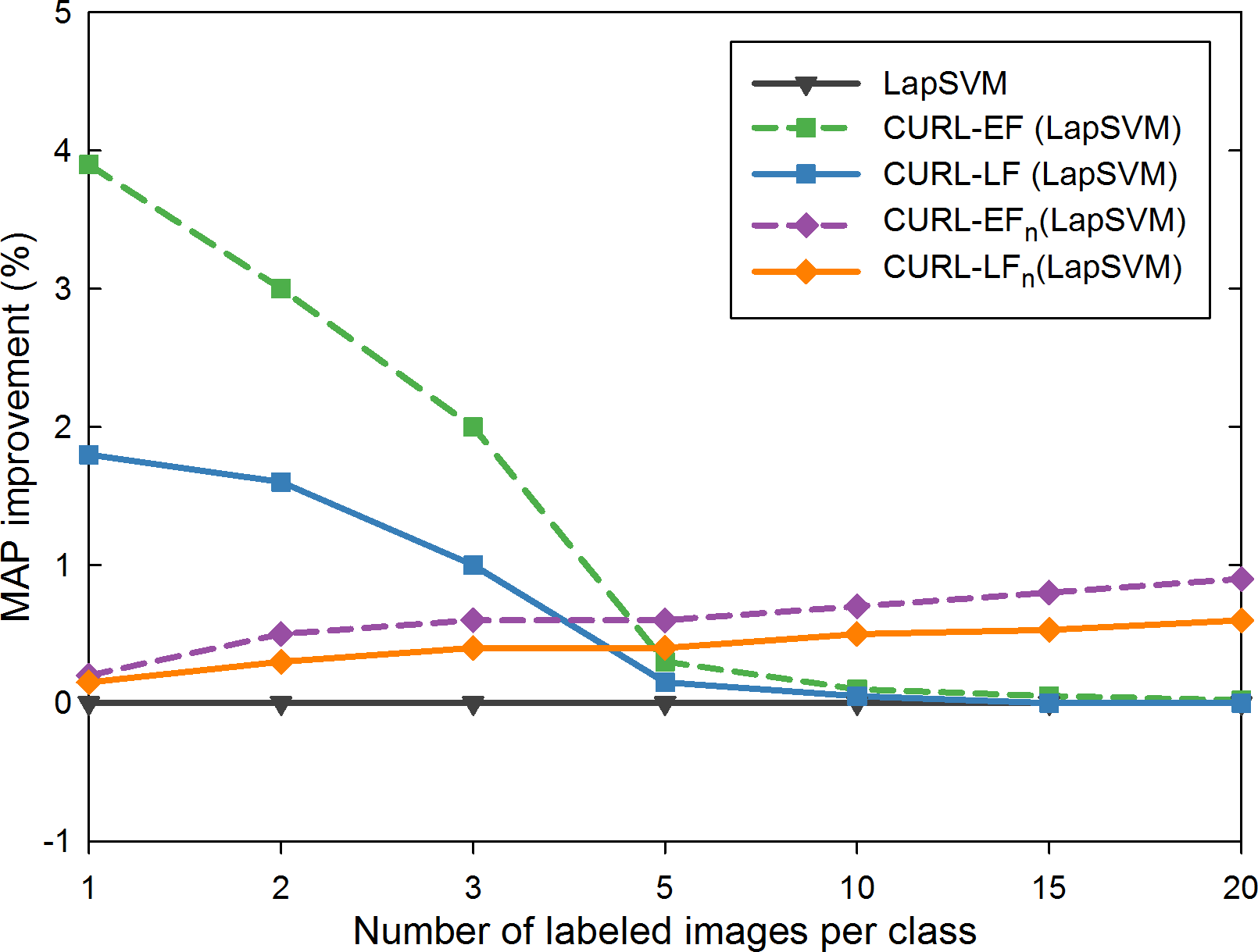} \\
\end{tabular}
\caption{Mean Average Precision (MAP) varying the number of labeled images per class, obtained on the ILSVRC 2012 data set: MAP values (left) and MAP improvements over LapSVM baseline (right). Results are obtained using GIST, PHOG, LBP and CNN features.}%
\label{fig:curllapin}%
\end{figure*}

The experimental results are reported in Fig. \ref{fig:curllapin} and represent the average performance over ten runs with random labeled-unlabeled feature splits. 

Given the large range of MAP values, the plot of MAP improvements with respect to LapSVM baseline is also reported. It can be seen that the behavior is similar to that of the previous plots, with the LapSVM-based CURL variants outperforming the LapSVM. As for the previous data sets, the plots show that \curllapef{ }and \curllaplf{ }are better suited for small sets of labeled images, while \curllapefn and \curllaplfn{ }are to be preferred when more labeled examples are available. It is remarkable that the proposed classification \coso{ }is able to improve the results of the LapSVM, since the CNN features were specifically learned for the ILSVRC 2012.

%
%

\section{Conclusions}
\label{sec:conclusions}
In this work we have proposed CURL, a semi-supervised image classification \coso{ }which exploits unlabeled data in two different ways: first two image representations are obtained by unsupervised learning; then co-training is used to enlarge the labeled training set of the corresponding classifiers. The two image representations are built using two different fusion schemes: early fusion and late fusion. 

The proposed \coso{ }has been tested on the Scene-15, Caltech-101, and ILSVRC 2012 data sets, and compared with other supervised and semi-supervised methods in three different experimental scenarios: inductive learning, transductive learning, and self-taught learning. We tested two {\emb}s of CURL and several variants differing in the co-trained classifier used and in the number of pseudo-labeled examples that are added at each co-training round. The experimental results showed that the CURL {\emb}s outperformed the other methods in the state of the art included in the comparisons. In particular, the variants that add a single pseudo-labeled example per class at each co-training round, resulted to perform best in the case of a small number of labeled images, while the variants adding more examples at each round obtained the best results when more labeled data are available.  

Moreover, the results of CURL using a combination of low/mid and high level features (i.e. LBP, PHOG, GIST, and CNN features) outperform those obtained on the same features by state of the art methods. This means that CURL is able to effectively leverage less discriminative features (i.e. LBP, PHOG, GIST) to boost the performance of more discriminative ones (i.e. CNN features). 



\ifCLASSOPTIONcaptionsoff
  \newpage
\fi


\bibliographystyle{IEEEtran}
\bibliography{cotrain}

%
%
%




\end{document}